\documentclass[review,3p,times]{elsarticle}

\usepackage{amsmath,amssymb}
\usepackage{enumitem}
\usepackage{footnote}

\usepackage{graphicx}
\usepackage[labelformat=simple]{subcaption}

\usepackage[colorlinks=true,urlcolor=blue,citecolor=blue]{hyperref}
\usepackage{afterpage}
\usepackage{geometry}
\usepackage{longtable}
\usepackage{pdflscape}
\usepackage{multirow}
\usepackage{array, makecell}
\usepackage{blindtext}

\usepackage{lineno}

\journal{Computer Science Review}

\begin{document}

\begin{frontmatter}



\title{A Review of Evolutionary Multi-objective Clustering Approaches}

\author[add1] {Cristina Y. Morimoto\corref{mycorrespondingauthor}}
\ead{cristina.morimoto@ufpr.br}

\author[add1] {Aurora Pozo}
\ead{aurora@inf.ufpr.br}

\author[add2] {Marc\'ilio C. P. de Souto}
\ead{marcilio.desouto@univ-orleans.fr}

\address[add1]{Department of Informatics, Federal University of Paran\'a, Curitiba-PR, Brazil}
\address[add2]{LIFO, University of Orl\'eans, Orl\'eans, France}
\cortext[mycorrespondingauthor]{Corresponding author}

\begin{abstract}
Evolutionary multi-objective clustering (EMOC), a modern clustering technique, has been widely applied to extract patterns, allowing us to analyze different aspects of complex data by considering multiple criteria. In this article, we present an analysis of the advances in EMOC studies and provide a profile of this study field by considering an extensive mapping of the literature to identify the main methods and concepts that have been adopted to design the EMOC approaches. This review provides a comprehensive view of the EMOC studies that supports newcomers or busy researchers in understanding the general features of the existing algorithms and guides the generation of new approaches. For that, we introduce a general architecture of EMOC to describe the main elements applied in designing EMOC algorithms and we correlate them with the main features found in the literature. Also, we categorized the EMOC algorithms based on shared characteristics that highlight the main features or application fields. The paper ends by addressing some potential subjects for future research.
\end{abstract}




\end{frontmatter}


\section{Introduction}
Clustering is a type of unsupervised learning whose goal is to find the underlying structure composed of clusters (groups or categories) in which objects or observations belonging to each cluster should share some relevant property (similarity) regarding the data domain.  Clustering analysis is widely adopted in different fields of application (e.g. marketing, medicine, bioinformatics) considering different research subjects, such as pattern analysis, decision making, data mining, and image segmentation~\cite{jain1999data,Liu2018}. 

In recent years, Multi-Objective Evolutionary Algorithms (MOEAs) have become one popular methodology for clustering~\cite{Liu2018}. Studies of multi-objective clustering have emerged and increased in the last two decades, exploring the use of multiple criteria to extract patterns and provide multiple partitions as solutions. Thus, some reviews and surveys have come out to present a general view of the features and applications of multi-objective clustering approaches. In 2009, Hruschka et al.~\cite{hruschka2009survey} introduced a general view of Evolutionary Multi-Objective Clustering (EMOC) in a review of evolutionary clustering algorithms. In 2012, Bong and Rajeswari~\cite{Bong2012} presented  multi-objective clustering trends and methods applied to image segmentation. In 2013, Mukhopadhyay et al.~\cite{ mukhopadhyay2013surveyb} presented a survey of multi-objective evolutionary approaches for data mining, in which the authors introduced the general features of the EMOC and some algorithms were presented. In 2015, Mukhopadhyay et al.~\cite{mukhopadhyay2015survey} introduced a basic framework and features of multi-objective clustering and a review of algorithms found in the literature was presented. However, the authors did not  provide the methodology applied to mapping  and selecting the algorithms presented in their work.  
In 2019, Gupta and Sharm~\cite{Gupta2019survey} presented a list of some algorithms focused on solving real-life problems. Finally, in 2021, Khurma and Aljarah~\cite{khurma2021review} provided a review that presented a general view of applications for multi-objective evolutionary clustering. However, in the entire review, only one EMOC algorithm was cited. 
In summary, each of these studies contributes to multi-objective clustering research by focusing on a specific scope or application field, but without taking into account a comprehensive mapping of existing EMOC approaches.

Aiming to provide a  broad view of the EMOC studies, this paper provides a review that considers a systematic mapping of articles in the ACM Digital Library, IEEE Xplore, and Scopus. We present the most relevant EMOC algorithms, considering high-impact papers based on the h-index and Scopus percentile scores. These algorithms were grouped by common features or strategies for data clustering.
To our knowledge, this is the first review of EMOC that presents how the studies in EMOC have evolved and the main topics associated with this research field. Furthermore, focusing on providing a guide for new practitioners and students of EMOC, we deal with all the components of the EMOC architecture to support the interested in implementation and assessment of EMOC algorithms. 

Moreover, our research goals are to identify and introduce the main features and research subjects of the existing EMOCs studies, highlighting the progression of the use of multi-objective algorithms and objective functions. 

The remainder of this paper is organized as follows. In Section~\ref{sec:preliminaries}, we present the main concepts and terms regarding clustering and multi-objective optimization, correlating these subjects. In Section~\ref{sec:MC}, we introduce a general architecture of evolutionary multi-objective clustering and compile the main strategies applied in the design of the EMOC approaches.  
Section~\ref{sec:analys} presents some numerical data on a set of multi-objective clustering studies to show the evolution of the publications of the EMOC. Then, in Section~\ref{sec:MCA}, we present a general review of the EMOC algorithms, considering algorithms for general purpose and specific applications.  Finally, Section~\ref{sec:remarks} highlights our main findings and discusses future works.

\section{Background}\label{sec:preliminaries}
In this section, we first introduce basic concepts in clustering and multi-objective optimization. We then describe general aspects of the assessment of multi-objective clustering.

\subsection{Clustering and Multi-objective Optimization}
Data clustering consists of the decomposition of finite and unlabeled data into subgroups based on similar attributes, or naturally occurring trends, patterns, or relationships in the data~\cite{Jain1988}. There is not a unique and formal definition of a cluster since the clustering methods and algorithms were proposed for researchers in different fields and applied to a variety of problems and distinct goals. In general, some general properties for cluster analysis are considered~\cite{hruschka2009survey,rai2010survey}:  
\begin{enumerate}[label=(\alph*)]
\item \textbf{Well-separated clusters} represent clusters where each object is closer (more similar) to all of the objects in its cluster than to any object in another cluster; \item \textbf{Connected or contiguous clusters} refer to clusters in which  each object is closer to at least one object in its cluster than to any object in another cluster;
\item \textbf{Compact clusters} represent clusters with small intra-cluster variation, considering the variation between same-cluster data items or between data items and clusters; 
\item \textbf{Center-based clusters} represent clusters in which  each object is closer to the center of its cluster than to the center of any other cluster;  
\item \textbf{Density-based clusters} denote clusters in which regions of high density are separated by regions of low density.
\end{enumerate}

In terms of the clustering process, in this paper, we consider two general types: hard and soft clustering. 
Formally,  given a set of objects  $ \mathbf{X} = \{\mathbf{x}_1, \mathbf{x}_2, \ldots, \mathbf{x}_n\} $, an hard (exclusive) partition of $X$ in $k$ clusters can be defined as $ \mathbf{\pi} = \{\mathbf{c}_1, \mathbf{c}_2, \ldots, \mathbf{c}_k \}$, where  $k < n$, such that: $\mathbf{c}_i \neq \emptyset$, for $(i = 1, \ldots, k)$, $\bigcup^{k}_{i=1} \mathbf{c}_i = \mathbf{X}$ and  $\mathbf{c}_i \cap \mathbf{c}_j = \emptyset$ for  $(i, j = 1, \ldots, k)$ and $i \neq j$. If the condition of mutual disjunction ($\mathbf{c}_i \cap \mathbf{c}_j = \emptyset$, for $(i, j =1, \ldots, k)$ and $i \neq j$) is relaxed, then the
corresponding data partitions are said to be of the soft (fuzzy) type \cite{hruschka2009survey}.

Regarding the taxonomy of the algorithms, traditional clustering algorithms can be divided into two general categories: partitional and hierarchical. Hierarchical methods produce a nested series of partitions, while partitional methods produce only one. For example, $k$-means (KM)~\cite{macqueen1967some} is a partitional algorithm; Single-Linkage (SL)~\cite{sneath1957}, Average-Linkage (AL)~\cite{sokal1958}, and Complete-Linkage (CL)~\cite{sorensen1948} are hierarchical algorithms.
In general, traditional clustering algorithms optimize only one clustering criterion and are often very effective for this purpose. However, they may not find all clusters in the datasets with different data structures, or clusters with shapes hidden in sub-spaces of the original feature space.

\begin{figure*}[htb]
  \begin{subfigure}{.32\textwidth}
    \centering
   \includegraphics[width=\textwidth]{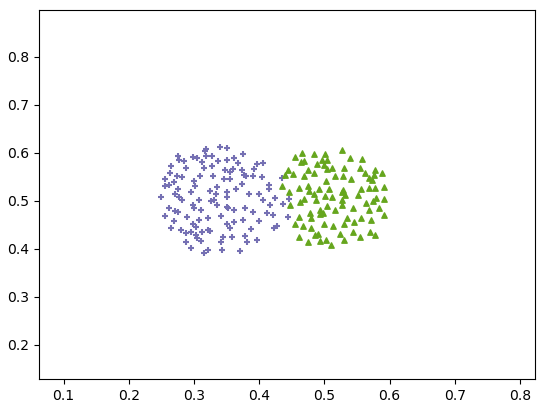}
    \caption{globular clusters}
    \label{sf1}
  \end{subfigure}
  \begin{subfigure}{.32\textwidth}
    \centering
    \includegraphics[width=\textwidth]{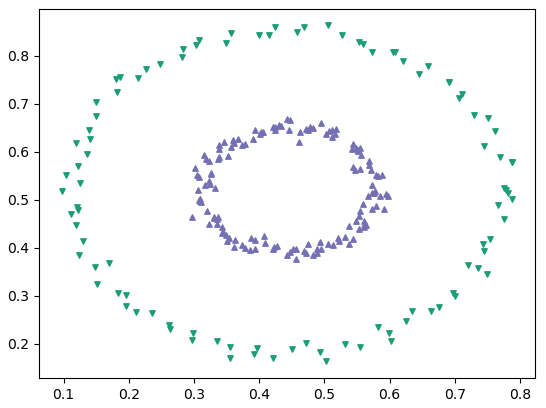}
    \caption{ring shape clusters}
    \label{sf2}
  \end{subfigure}
  \begin{subfigure}{.32\textwidth}
    \centering
    \includegraphics[width=\textwidth]{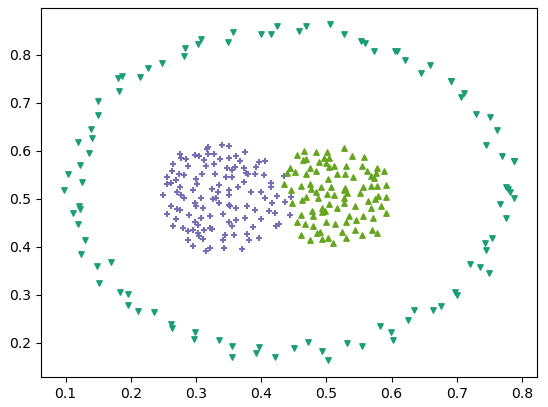}
    \caption{heterogeneous data structure}
    \label{sf3}
  \end{subfigure}
  \caption{Different data structures (objects with the same color represent a cluster in each sub-figure)}
  \label{fig_st}
\end{figure*}

In contrast, Evolutionary Multi-Objective Clustering (EMOC), a modern clustering type of algorithm, considers the simultaneous optimization of multiple objectives to solve a variety of clustering problems considering different data properties.
An EMOC that considers two criteria, compactness-based and connectedness-based, for example, can detect all of the data structures in Fig. \ref{fig_st}, whereas algorithms that use only the compactness-based criterion, such as $k$-means (KM)~\cite{macqueen1967some}, can detect globular clusters, as shown in Fig.~\ref{sf1}, but KM cannot find the  ring-shaped clusters in Fig.~\ref{sf2} and the heterogeneous structures in Fig.~\ref{sf3}. In contrast, a connectedness-based algorithm, such as Shared Nearest Neighbor (SNN)~\cite{ertoz02}, can detect the ring shapes in Fig.~\ref{sf2}, but SNN cannot find the clusters in  Fig.~\ref{sf1} and  Fig.~\ref{sf3}.

EMOC  applies the concepts of multi-objective optimization (MOO) to the clustering problem. In MOO, the goal is to find a vector of decision variables, $\mathbf{\pi}$,  that satisfies the inequality and equality constraints ($g_i(\mathbf{\pi}) \le 0, \quad i=\{1, \ldots, p\}, \textrm{and }  h_j(\mathbf{\pi})=0, \quad j=\{1,  \ldots, q\}$), and optimizes the vector $\mathbf{F(\pi)}$ of $z$ objective functions, Eq. (\ref{mop})~\cite{Li2015}.
In other words, the Multi-objective Optimization Problem (MOP) involves the minimization (or maximization) of the vector function $\mathbf{F(\pi)}$, mapping a tuple of  parameter decision variables to a tuple of objectives, where $z \geq 2$~\cite{Zitzler1999}.

\begin{equation}\label{mop}
\textrm{minimize/maximize } \mathbf{F}(\mathbf{\pi}) = (f_1(\mathbf{\pi}), \ldots, f_{z}(\mathbf{\pi}))
\end{equation}

\begin{figure}[ht]
  \centering
  \includegraphics[width=0.48\textwidth]{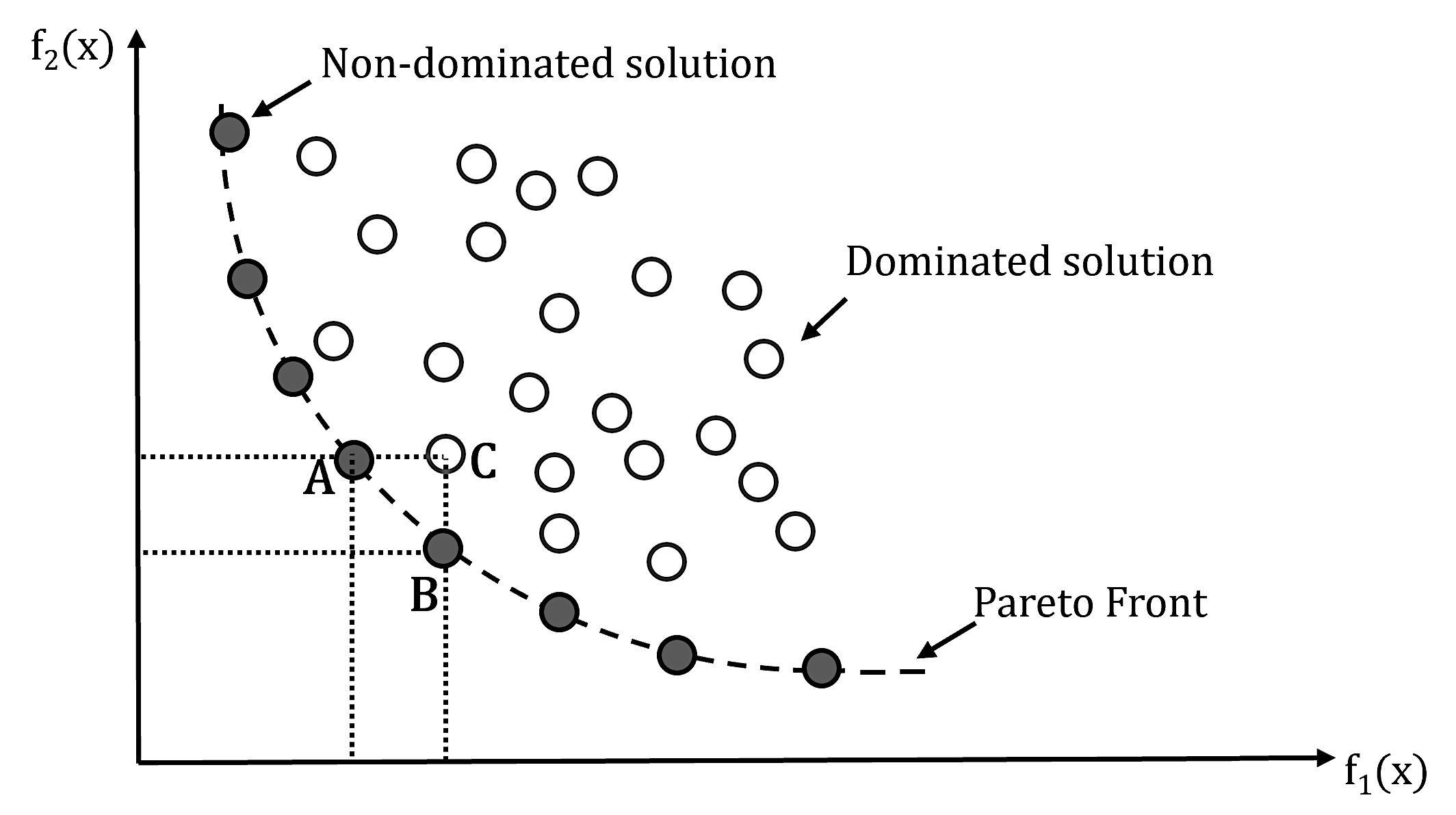}
  \caption{Pareto Dominance Relation}
  \label{fig:dominance}
\end{figure}

Evolutionary Algorithms (EAs) are considered well-suitable to MOO because they address both search and multi-objective decision making (while some approaches focus on search and others on multi-criteria decision making) and can search partially ordered spaces for several alternative trade-offs~\cite{Fonseca1995}.
EA uses a heuristic solution-search or optimization technique based on the principle of evolution through selection. Most multi-objective evolutionary algorithms select solutions using the Pareto dominance relation, in which given two candidate solutions $\pi_i$ and $\pi_j, \pi_i$ dominates $\pi_j$ (denoted as $ \mathbf{\pi}_i \prec \mathbf{\pi}_j$), if and only if: i) $\pi_i$  is strictly better than $\pi_j$ in at least one of all the objectives considered, and ii) $\pi_i$ is not worse than $\pi_j$ in any of the objectives considered. The goal of this process is to find the set of all non-dominated solutions, that is, the Pareto-optimal front (PF). For example, Fig.~\ref{fig:dominance} shows a Pareto set of two objective functions that should be minimized. Points A and B are the non-dominated solutions and hence lie on the Pareto front. Point C is dominated by points A and B, so it does not lie on the frontier\cite{Li2015}. 

Due to their population-based nature, evolutionary algorithms are able to approximate the whole PF of a given multi-objective problem in a single run. Consequently, they have been a popular choice for the design of multi-objective data clustering techniques~\cite{hruschka2009survey, mukhopadhyay2015survey}. In this context, the Multi-objective Evolutionary Algorithms (MOEA) are applied to solve a MOP  with $z \geq 2$. However, the traditional techniques based on Pareto dominance have their effectiveness degraded (convergence and diversity difficulties) when applied to problems with more than three objectives, and the computational complexity of non-dominated sorting considerably increases. Many-Objective Evolutionary Algorithms (MaOEA) have been proposed to deal with this scalability issue, in which the Many Objectives Problem can be defined as a MOP with $z \geq 4$~\cite{Li2015}.

In terms of the evaluation of the EMOC results, there are two types of assessment: one considering aspects of clustering quality, and the other considering MOO performance, as presented in the following.

\subsection{Clustering Validation}\label{sec:cvis}
The clustering approaches are evaluated regarding  Clustering Validity Indices (CVIs), which define how well a partition fits the structure underlying the data. There are three types of criteria~\cite{Brun2007}: relative, internal, and external. 
{Relative criteria} are based on comparisons of partitions generated by the same algorithm with different parameters or different subsets of the data. {Internal criteria} refer to quality measures based on calculating properties of the resulting clusters, establishing the validity of a cluster-based exclusively on the dataset itself, for example, how much a cluster is justified by means of the proximity matrix.  {External criteria} lie in prior knowledge of structures in the dataset to evaluate the given partitions generated by an algorithm in contrast with a model partition or labeled data, denominated True Partition\footnote{The True Partition or ground truth is the labeled data that form the real partition, the underlying structure of the data. }, provided by specialists. In Section~\ref{sec:obj_fc}, we detail the CVIs and their application in EMOC approaches.

\subsection{Performance in multi-objective optimization}\label{sec:emocP}
There are a variety of quality indicators applied to MOO, as presented in~\cite{riquelme2015performance}. These indicators are used to determine the convergence and diversity of the solution.
Convergence is to measure the ability to  attain a global Pareto front, and diversity is to measure the distribution along the Pareto front. Here, we introduce two popular indicators, IGD - {Inverted Generational Distance} and  HV - {Hypervolume}.

The IGD index  is computed in the objective space, which can be viewed as an approximate distance from the Pareto front to the solution set in the objective space. So, given a set of solutions $\mathbf{S}$ and a set of $\mathbf{R}$ uniformly distributed representative points of the PF, the IGD measure is computed according to Eq.~\ref{eq:igd}, where $d(\mathbf{r},\mathbf{S})$ is the minimum Euclidean distance between $\mathbf{r}$ and the points in $\mathbf{S}$, and $\lvert R\rvert$ is the cardinality of $R$. A lower IGD result refers to a better quality of S~\cite{zhang2018mullti}.

\begin{equation}
    IGD(\mathbf{S}, \mathbf{R}) = \frac{\sum_{\mathbf{r} \in \mathbf{R}} \min\{d(\mathbf{r},\mathbf{S})\}}{\lvert R\rvert}
	\label{eq:igd}
\end{equation}

HV measures the volume of the area enclosed by the set and a reference point specified by the user.
The hypervolume formula is given by Eq.~\ref{eq:hv}, where $vol$ refers to the Lebesgue measure and $\mathbf{z}=\{\mathbf{z}_1, \ldots, \mathbf{z}_m\}$ is a given reference point, the nadir point $\mathbf{z}_{nad}$. A nadir point corresponds to the worst Pareto-optimal solution of each objective, and the nadir objective vector represents the worst value of each objective function corresponding to the entire Pareto-optimal set~\cite{zitzler1998multiobjective}.

\begin{equation}
	HV(\mathbf{S}) = vol(\bigcup_{\mathbf{x} \in \mathbf{S}} [f_1(\mathbf{x}), \mathbf{z}_1] \cdot  \ldots \cdot  [f_m(\mathbf{x}), \mathbf{z}_m])
	\label{eq:hv}
\end{equation}

The HV metric  reflects the solutions’ quality in terms of both convergence and maximum spread. A larger HV value indicates a better approximation to the PF.

\section{A General Architecture of Evolutionary Multi-objective Clustering}\label{sec:MC}

\begin{figure*}[ht]
  \centering
  \includegraphics[width=\textwidth]{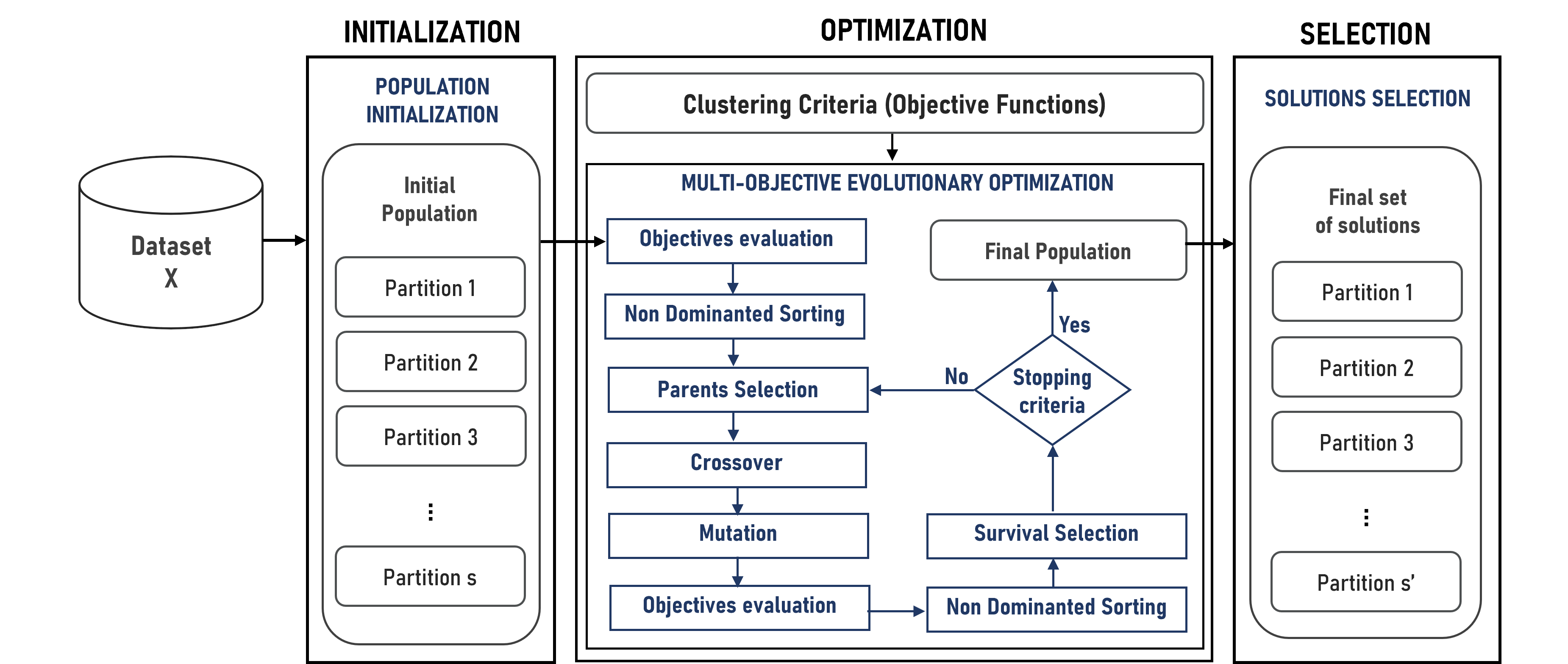}
  \caption{A general architecture of Evolutionary Multi-objective Clustering }
  \label{fig:architecture}
\end{figure*}

In this section, we present the general architecture of an evolutionary MOO, considering the clustering problem. We illustrate the general components of the EMOC in Fig.~\ref{fig:architecture} considering 3 modules: 
\begin{enumerate}
	\item \textbf{Initialization}: Given a dataset, traditional data clustering algorithms (or random generator methods) are applied to build the partitions (individuals) that compose the initial population. Each partition is a clustering solution with a specific encoding or representation. 
    In section \ref{sec:emoc_ini}, we detail the types of representations and initialization strategies applied in EMOC. 
	\item \textbf{Optimization}: The initial population is taken as an input to multi-objective evolutionary optimization, in which iteratively the objective functions are minimized (or maximized) to generate a final population.	
    In general, the existing EMOC algorithms rely on general-purpose MOEAs in the optimization flow, in which most approaches consider the standard features of a particular MOEA, while using a specific set of objective functions and different combinations of crossover and mutation operators.
    In section \ref{sec:emoc_moea}, we detail the optimization phase, in which we present some traditional MOEAs and introduce other types of multi-objective approaches that consider other aspects in the selection besides Pareto dominance. Furthermore, we point out the main aspects of the objective functions and the evolutionary operators applied in EMOC.     
	\item \textbf{Selection}: MOO approaches may generate large sets of efficient solutions using Pareto
    dominance. Thus, this module is applied to determine the final set of solutions to be presented to the data experts. According to prior criteria, a suitable number of solutions, $s^\prime$, is selected from the final population in this phase. The partition selection is a specific subject in clustering, in which it is  possible to find studies focused on this subject. Therefore, this module is not considered mandatory in the design of EMOC approaches. In 
	Section \ref{sec:emoc_sel}, we present some strategies applied to EMOC partition selection.
\end{enumerate}

In the following, we present  the main concepts and elements of each module of evolutionary multi-objective clustering by introducing the main features of the EMOC approaches described in Section~\ref{sec:MCA}. 

\subsection{Initialization Module: Representation and Initialization strategies}\label{sec:emoc_ini}
The solution representation or chromosome encoding denotes an individual in the evolutionary algorithm. The choice of the representation should consider the information necessary to be manipulated by the evolutionary operators to generate new feasible solutions. In general, the most popular types of clustering representation solutions for EMOC are~\cite{hruschka2009survey}: 
\begin{enumerate}[label=(\alph*)]
    \item \textbf{Label-based} representation, which takes into account labels for each object in the partition. The length of an encoding of the solution is equal to the number of objects in the dataset, and each position denotes the cluster label of the respective object. 
    \item \textbf{Prototype-based} encoding  is usually applied in centered-based clustering, in which cluster prototypes, such as centroids, medoids, or modes, are used in partition representation. In the centroid-based encoding, the chromosomes are denoted by the coordinates of the cluster centers. In medoid-based encoding, the chromosomes are represented by the coordinates that define the smallest average dissimilarity of the cluster to all other objects. In mode-based encoding, the chromosome can denote the frequency of the attribute. In general, in the prototype-based representation, one can have $k$ chosen centers, in which the objects at each point are associated with the closest chosen center measure. 
    \item \textbf{Locus-based adjacency graph} (LAG) representation corresponds to a graph containing a vertex for each data point, and the links between two data points represent the edges. The linked objects represent the clusters in the solution.
\end{enumerate}

In particular, some approaches use a binary representation to define the labels or prototypes instead of using numerical values. In \cite{Sert2011, sert2012unification}, each chromosome includes $n \cdot k$ bits, and each reserved $k$ bits  provides the cluster number of the corresponding instance. In ~\cite{Ripon2009}, each data point is a candidate center, and a binary encoding is applied to define whether a data point is a center or not.
Besides that, it is possible to consider other aspects of the clustering problem in the representation. For example, in \cite{DiNuovo2007}, FCM parameters and feature weights are applied to represent the solution. In \cite{Zhu2012, xia2013novel, zhou2018kernel}, they use the center information associated with a center weight to encode the solutions. In \cite{dong2018adaptive, Zhu2018_MaOEA}, the fuzzy membership matrix and the center information are designed to represent each solution. In \cite{Luo2015}, they consider an input as a linear combination of base elements (e.g., parameters or coefficients), which are chosen from an over-complete dictionary to design the sparse-based representation.

Regarding the initialization, a common practice in EMOC approaches is to use random generators to assign labels or choose the initial centers of the clusters in the partition.
The random initialization generally provides unfavorable partitions since the clusters are likely to be mixed up to a high degree. However, this strategy is very popular because of its simplicity and effectiveness in testing the algorithms against hard evaluation scenarios~\cite{hruschka2009survey}.

In contrast, some relevant EMOC algorithms use high-quality  individuals in the initial population, in which  clustering algorithms are applied to generate the base partitions. For example, KM, AL, SL, CL, Minimum Spanning Tree (MST) - clustering~\cite{Handl2005a}, Shared Nearest Neighbor (SNN)~\cite{ertoz02}, Spectral Clustering (SPC)~\cite{shi2000} are applied in the initialization  of some EMOCs presented in Section~\ref{sec:MCA}.   

In the literature, most prototype-based encoding approaches use random generators in the initialization. On the other hand, the label-based encoding takes advantage of not requiring  decoding of the solutions,  making it possible to apply most of the traditional clustering algorithms in the initialization.
The LAG representation can rely on a graph-based method in the initialization, such as MST-clustering, taking advantage of its data structure.

\subsection{Optimization Module: Multi-objective Evolutionary Optimization}\label{sec:emoc_moea}
In general, the EMOC algorithms rely on general-purpose MOEAs in the optimization module. 
The choice of the multi-objective approach should consider the number of objective functions and the characteristics of the application, in which it is possible to explore some aspects, such as user preference, diversity of solutions, among other features.

The most traditional category of multi-objective algorithms is \textbf{Pareto-based}, where the solutions are evaluated and compared by considering the Pareto dominance. For example, the NPGA - Niched Pareto Genetic Algorithm~\cite{Horn1994} is designed along with the natural analogy of the evolution of distinct species exploiting different niches or resources in the environment, in which the main strategy  relies on tournament selection among a population's individuals and Pareto dominance. The PESA-II - Pareto Envelop-based Selection Algorithm version 2~\cite{Corne2000} is an elitist method (the selection considers the best one or more solutions, called the elites, in each generation,  which are inserted into the next), where the diversity mechanism is cell-based density. The NSGA-II - Non-dominated Sorting Genetic Algorithm version II~\cite{Deb2002} is an elitism method that employs a ranking based on non-domination sorting associated with crowding distance. The SPEA-2 - Strength Pareto Evolutionary Algorithm version 2~\cite{Zitzler2001} is also an elitism method that applies the concept of the strength of dominators as a fitness assignment, employing a density based on the $k$th nearest neighbor to preserve the diversity. 

Beyond that, Li et al.~\cite{Li2015} defined other categories by considering other aspects beyond the Pareto front to evaluate and compare the solutions in MOEAs/MaOEAs: 
\begin{enumerate}[label=(\alph*)]
    \item \textbf{Relaxed dominance-based} algorithms use a variant of dominance, such as value-based (that  changes the objective values by modifying the Pareto dominance of the solutions when comparing them) or  number-based dominance (that compares a solution to another by counting the number of objectives where it is better than, the same as, or worse than the other);
    \item  \textbf{Diversity-based} algorithms apply a customized diversity-based approach, for example, the SDE (Shift-Based Density Estimation), where the diversity is taken as the first criterion instead of the convergence; it is possible because SDE shifts the positions of the solutions to measure the density of the neighborhood of the solution, allowing both the distribution and the convergence information to be used in the comparison of the solutions;
    \item  \textbf{Aggregation-based} algorithms apply aggregation functions to evaluate the solutions, which can be divided into two categories: aggregation of objective values and aggregation of objective ranks.
    \item  \textbf{Indicator-based} algorithms aim to maximize the value of a specific indicator, which can be divided into three classes: hypervolume driven, distance-based indicator driven, and R2 indicator driven; 
    \item  \textbf{Preference set-based} algorithms consider the user’s preferences in the optimization process. This kind of algorithms can be divided into three classes based on the timing of the set of preferences being used: a priori (selection before the search), interactive (selection during the search), and a posteriori (selection after the search); 
    \item  \textbf{Reference-based} algorithms consider a set of reference solutions, which are applied to measure the quality of the solutions and guide the search during the evolutionary optimization process, such as in NSGA-III~\cite{Deb2014} and  RVEA~\cite{Cheng2016};
    \item  \textbf{Dimensionality reduction} algorithm seeks to simplify the problem by reducing its complexity,  where the number of objectives can be reduced gradually during the search process (online) or the dimensionality reduction is carried out after obtaining a set of Pareto-optimal solutions (offline).
\end{enumerate}
Additionally, it is possible to consider another category, a \textbf{Hybrid-based}, that  combines two or more approaches to overcome their particular problems, for example, the MOEA/DD - {Multi-Objective Evolutionary Algorithm based on Dominance and Decomposition approaches}~\cite{Li2015b} combines two categories of strategies: Pareto dominance and aggregation.

As mentioned above, in general, MOEAs are applied to clustering problems, considering specific objective functions (clustering criteria), and different combinations of crossover and mutation operators. Thus, we detail them in the following sub-sections.

\subsubsection{Objective Functions}\label{sec:obj_fc}
In general, CVIs (see Section \ref{sec:cvis}) that consider internal and relative criteria are used as clustering objective functions. On the other hand, specific objective functions designed for multi-objective clustering, such as the sparsity ($SP$) and reconstruction error ($RE$) designed for spectral clustering, can be used in EMOCs~\cite{Luo2015}. 

In the following, we introduce objective functions categorized by criteria (cluster properties). These objective functions denote the clustering criteria adopted in the EMOCs presented in Section \ref{sec:MCA}: 
\begin{enumerate}[label=(\alph*)]
    \item \textbf{Compactness criteria}: average within group sum of squares ($AWGSS$)~\cite{Kirkland2011},  overall deviation ($Dev$)~\cite{Handl2005a}, K-Mode internal distance ($Km_{id}$)~\cite{Sert2011}, K-Mode weighted internal distance ($Km_{wid}$)~\cite{Sert2011}, intra-cluster entropy ($Ent$)~\cite{Ripon2006_ICPR}, homogeneity ($H$)~\cite{Dutta2012}, intra-cluster variance ($Var$)~\cite{Garza2018}, and total within-cluster variance ($TWCV$)~\cite{Du2005}, and fuzzy compactness ($J_m$)~\cite{bezdek2013pattern}, are criteria based on intra-cluster similiarity.

    \item \textbf{Connectedness criteria}: connectivity index ($Con$)~\cite{Handl2005a}, and data continuity degree ($DCD$)~\cite{menendez2013multi}, are criteria based on neighborhood relationship.   

    \item \textbf{Separation criteria}:  average between-group sum of squares ($ABGSS$)~\cite{Kirkland2011}, inter-cluster average separation ($Sep_{AL}$)~\cite{Ripon2009}, K-Mode external distance ($Km_{ed}$)~\cite{Sert2011}, K-Mode weighted external distance ($Km_{wed}$)~\cite{Sert2011}, separation index ($Sep_{CL}$)~\cite{dutta2012data}, and graph-based separation ($Sep_{graph}$)~\cite{menendez2013multi}, fuzzy separation ($Sep_{fuzzy}$)~\cite{Mukhopadhyay2007icit}, and fuzzy overlap separation ($Sep_{nfuzzy}$)~\cite{Wikaisuksakul2014}, are criteria based on inter-cluster similarity.  

    \item \textbf{Separation and Compactness criteria}: categorical data clustering with subjective factors ($CDCS$)~\cite{Zhu2018_MaOEA}, Calinski-Harabasz ($CH$)~\cite{Zhu2018_MaOEA}, Davies-Bouldin ($DB$)~\cite{Zhu2018_MaOEA}, Dunn~\cite{dutta2019automatic}, modularity ($Mod$)~\cite{Liu2018}, silhouette ($Sil$)~\cite{Mukhopadhyay2007}, $\mathcal{I}$~\cite{dong2018adaptive}, addition feature weight ($J_{Add}$)~\cite{xia2013novel}, Xeni-Beni ($XB$)~\cite{DiNuovo2007}, soft subspace Xie-Beni ($SSBX$)~\cite{Zhu2012}, are criteria that take into account both intra-cluster and inter-cluster similarity.

    \item \textbf{Other criteria}: cluster cardinality index ($CCI$)~\cite{Zhu2018_MaOEA} and expected weighted coverage density ($EWCD$)~\cite{Sert2011} consider the relation of the occurrence of the objects in a categorical dataset. The similarity index ($Sim$)~\cite{li2017ensemble} is the only relative CVI that compares partitions used as the objective function, while the other CVIs consider the data properties of each partition.  
\end{enumerate}

It is a common practice in the literature to apply 2 or more different  categories of clustering criteria as objective functions, where the approach will be able to optimize multiple characteristics of the evolved clusters. For example, a popular pair of objective functions, ($Var$, $Con$), consider  the compactness and  connectedness criteria.
In Section \ref{sec:MCA} other combinations of objective functions are presented. 
Due to the large number of clustering criteria and considering that some objective functions may have different names in the literature, we present a detailed description of each of these objective functions in \ref{appA}.

\subsubsection{Crossover and Mutation Operators}\label{sec:operators}
Evolutionary optimization relies on crossover and mutation operators to generate new solutions. In the literature,  we can find approaches using traditional evolutionary operators and clustering designed operators. The most popular traditional operators used in EMOC approaches are:
\begin{enumerate}[label=(\alph*)]
    \item \textbf{One-Point crossover}: one crossover point is considered along the length of the parents' chromosomes, and the genes following the crossover point in  one parent are swapped with the genes in the other parent~\cite{hruschka2009survey}. 
    \item \textbf{Two-Point crossover}: two crossover points along the length of the chromosome of each parent, such that the interval of genes between these two points are swapped~\cite{hruschka2009survey}.
    \item \textbf{Shuffle crossover}: this operator is similar to one-point crossover, in which a single crossover position is selected, and before the variables are exchanged, they are randomly shuffled in both parents~\cite{Sert2011}.
    \item \textbf{Uniform crossover}: for each position on the chromosome, a random decision is made on whether the swapping of genes should be done or not~\cite{Handl2007}.
    \item \textbf{Simulated binary crossover (SBX)}: this operator uses a probability density function that simulates the One-Point Crossover in binary-coded representation~\cite{Wikaisuksakul2014}.  
    \item \textbf{Polynomial mutation}: a polynomial probability distribution is applied to perturb a solution~\cite{Ripon2006_IJCNN}. 
    \item \textbf{Uniform mutation}: this operator replaces the value of the chosen particular slot position with a uniform random value selected considering a specified upper and lower bounds for that position~\cite{dong2018adaptive}.
\end{enumerate}

In terms of the clustering-designed operators, the representation and clustering criteria are taken into consideration. For example, the perturbation or replacement of center, centroid, or medoid is applied in the algorithms that use a prototype-based encoding to shift a randomly selected center slightly from its current position or replace the position of the cluster prototype according to a criterion; the exchange of the prototypes considers two parents in which there is an exchange of centroids to generate a new solution. Also, there are operators designed to split the objects of a cluster or merge two or more clusters to generate new solutions. 
Handl and Knowles~\cite{Handl2005b} presented the neighborhood-based mutation that is applied to the graph-based representation, replacing an existing link in the graph with another link to one of the randomly selected nearest neighbors. 
In \cite{bousselmi2017bi, bechikh2019hybrid}, Cheng and Church's (CC) algorithm was adapted to be applied as a mutation operator. The CC algorithm considers three steps (multiple node deletion, single node deletion, and node addition) to iteratively perform the removal and addition of rows and columns in a data expression matrix. As a mutation operator, only row operations are performed to preserve specific data properties.
Besides that, Faceli et al.~\cite{Faceli2006} introduced the use of clustering ensembles as a crossover operator. A clustering ensemble is a technique applied to combining multiple different clustering results (generated by different clustering algorithms or the same algorithm with different iterations) into a single partition~\cite{boongoen2018cluster}. As a crossover operator, pairs of partitions are combined with a consensus function to generate new individuals.

\subsection{Selection Module: Selection of Final Solutions}\label{sec:emoc_sel}
The Selection module is applied to restrict the number of clustering solutions presented to the decision-maker or data specialist. In the literature, most EMOCs select the final set of solutions by applying CVIs (see Section \ref{sec:cvis}). For example, in \cite{Tsai2012}, Pakhira, Bandyopadhyay and Maulik ($PBM$) ~\cite{Pakhira2004}, and $DB$ indices were used to single out the optimal solution. In \cite{menendez2013multi, menendez2014co}, the solution with the highest value of the $Sep_{graph}$ in the Pareto front was considered the best solution to be selected. In \cite{xia2013novel}, a new indicator called the projection similarity validity index (PSVIndex) was designed to select the best solution and cluster number. In \cite{dutta2019automatic}, the EMOC approach uses an overall rank of nine CVIs to determine the final set of solutions: C index~\cite{baker1976graph}, COSEC - Compactness and Separation Measure of Clusters~\cite{RAHMAN2014345}, $DB$, $Dunn$, $Dev$, $Ent$, $XB$, Purity~\cite{schutze2008introduction} and F-Measure~\cite{Larsen1999}. In particular, in \cite{Luo2015}, the non-dominated solutions are used to construct a standard adjacency matrix, and the measurement Ratio Cut \cite{Wei1999} provides a way to select a final trade-off solution.

Another way to select final solutions is by applying the knee-based approaches that are usually applied in determining the number of clusters in a data set. For example, the knee method presented by Handl and Knowles~\cite{Handl2005b} compares the final set of solutions and a control front. The solution corresponding to the largest distance between the actual non-dominated front and the control fronts is chosen to be the final solution, corresponding to the "knee" (the point of inflection) of the non-dominated front. 
In ~\cite{wang2018, Du2005}, the best clustering result is defined by the "elbow" method, which consists of picking the "elbow" or "knee" of the curve in the non-dominated front.

Besides that, clustering ensemble methods are used to select the final solutions. The non-dominated solutions are used as base partitions to generate the consensual partition by applying a consensual function to combine the base partitions.

\subsection{Evaluation of the EMOC algorithms} \label{sec:moec_evalutation}
In terms of evaluating clustering results, most EMOC approaches consider an external validity index, such as the adjusted Rand index (ARI)~\cite{Rand1971}, to evaluate the set of final solutions. ARI is  a corrected-for-chance version of the Rand index~\cite{Hubert1985}, computes the probability of two objects  of two partitions belong to the same cluster or different clusters, as defined in Equation (\ref{ARI}), where $n_{ij}$ is the number of common objects between the clusters $\mathbf{c}_i$ in $\pi_a$ and $\mathbf{c}_j$ in $\pi_b$, $n_i$ is the number of objects in the cluster $\mathbf{c}_i$ in  $p_a$, e   $n_j$ is the number of objects in the cluster $\mathbf{c}_j$ in $\pi_b$,  $k_a$ and $k_b$ are the number of clusters in the partitions  $\pi_a$ and  $\pi_b$.

\begin{equation}\label{ARI}
ARI =\frac{\sum\limits^{k_a}_{i=1}\sum\limits^{k_b}_{j=1} \binom{n_{ij}}{2} - 
    \left[\sum\limits^{k_a}_{i=1}\binom{n_{i}}{2} \cdot\sum\limits^{k_b}_{i=1}\binom{n_{j}}{2} \right] / \binom{n}{2}}{
    \frac{1}{2}\cdot{\left[\sum\limits^{k_a}_{i=1}\binom{n_{i}}{2} + \sum\limits^{k_b}_{i=1}\binom{n_{j}}{2} \right]} -{\left[\sum\limits^{k_a}_{i=1}\binom{n_{i}}{2} \cdot \sum\limits^{k_b}_{i=1}\binom{n_{j}}{2} \right]}/ \binom{n}{2}}
\end{equation} 

Besides that, the analysis of internal criteria can also be applied to investigate specific data structures. For example, in \cite{Ripon2006_ICPR, Ripon2006_IJCNN}, $H$, $Sep_AL$, $Dunn$, and $Dev$ are evaluated to analyze the general behavior of the EMOC approaches regarding each criterion. In \cite{dutta2012data, dutta2012simultaneous}, they compare their approaches with other ones based on the $DB$, $H$, and $Sep_{AL}$. 

It is important to observe that, rather than only  using the CVIs (See Section \ref{sec:cvis}) to analyze the algorithm performance, the evaluation of the optimizer can generate essential information  regarding the modeled problem. The quality indicators of the multi-objective optimization (see Section \ref{sec:emocP}) measure how well the final population reaches the goal of obtaining a converging and diverse set of solutions compared to the initial population.

\section{Overview of Multi-objective Clustering Studies}\label{sec:analys}
In this section, we show data analysis on a set of Multi-Objective Clustering (MOC) studies. 
This survey considers papers related to MOC from IEEE Xplore\footnote{\url{https://ieeexplore.ieee.org}}, ACM Digital Library\footnote{\url{https://dl.acm.org/}} and Scopus\footnote{\url{https://www.scopus.com}}. These article repositories contain the most important journal papers and conference proceedings, in the computer science and engineering domains.
We used the terms "multi-objective", "multi-objective", and "many-objective" as keywords related to optimization with multiple objectives, along with the term "clustering" to search by title for articles about multi-objective clustering. The article mapping considered English-language papers that were published before the year 2021.
The search result is 231 papers from IEEE Xplore, 30 papers from the ACM Digital Library, and 533 papers from Scopus, totaling 794 papers. Then, duplicated papers were removed. Finally, we analyzed the main contents of the resulting set of documents,  concluding with 358 papers. 
In the following, we discuss statistics on the publication of MOC works.

\begin{figure}[ht]
  \centering
  \includegraphics[width=\textwidth]{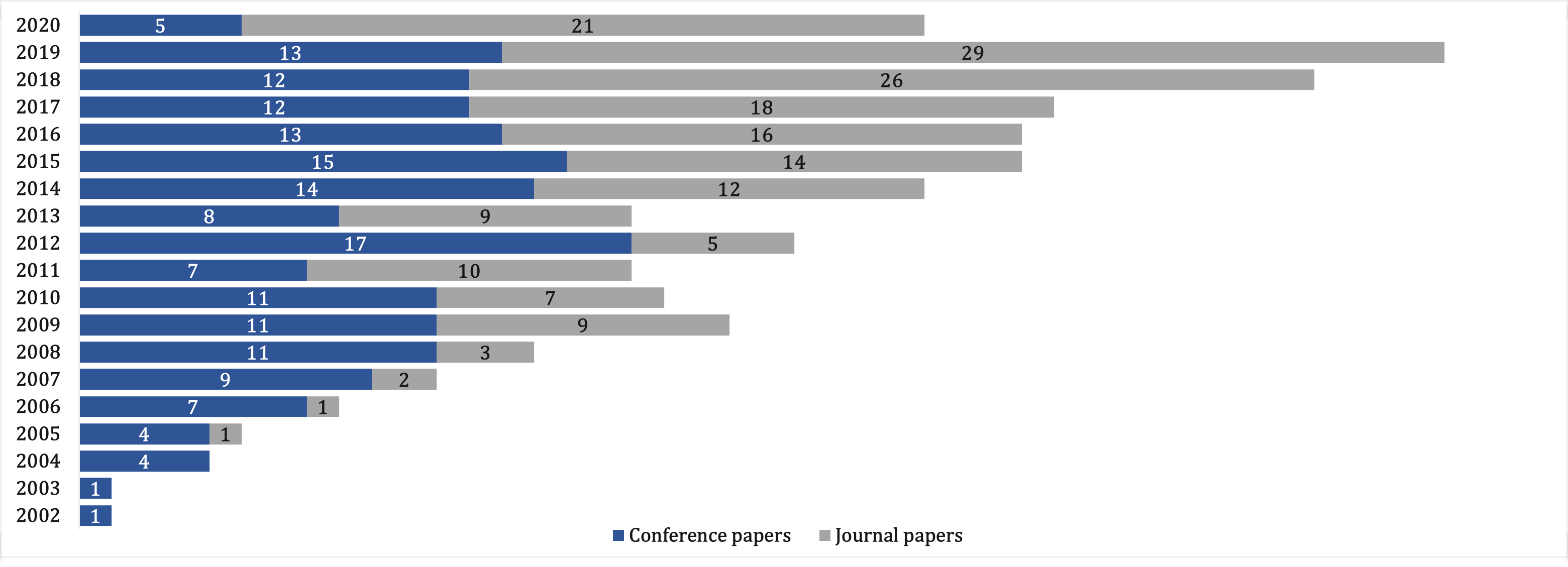}
  \caption{The number of publications related to MOC from 2002 to 2020.}
  \label{fig:articles}
\end{figure}

Fig.~\ref{fig:articles} shows the number of publications related to MOC that appeared in both journals and conferences over the years. It provides information on how the MOC field is evolving, based on the number of papers published. 
The first indexed article found was published in 2002~\cite{Zwir200265}, a conference paper in the Annals of the New York Academy of Sciences.  
In the same way, most of the articles published between 2002 and 2008 were published at conferences. In 2009, we observed a substantial increase in journal papers. Between 2008 and 2016, we verified a certain equilibrium in the number of articles published in conferences and journals, except  in 2012, when the number of conference papers increased abnormally, without a specific explanation.  
Finally,  in the last four years (2017-2020), the number of articles published in journals substantially increased. In particular, in 2019, the number of publications in journals was almost three times greater than the number of papers presented at conferences. In 2020, we can notice that the total number of papers significantly decreased compared to 2018 and 2019. One reasonable motivation was the COVID pandemic, which  motivated periods of suspension of non essential activities, and some conferences worldwide were canceled or postponed.

\begin{figure}[!htb]
  \centering
  \includegraphics[width=\textwidth]{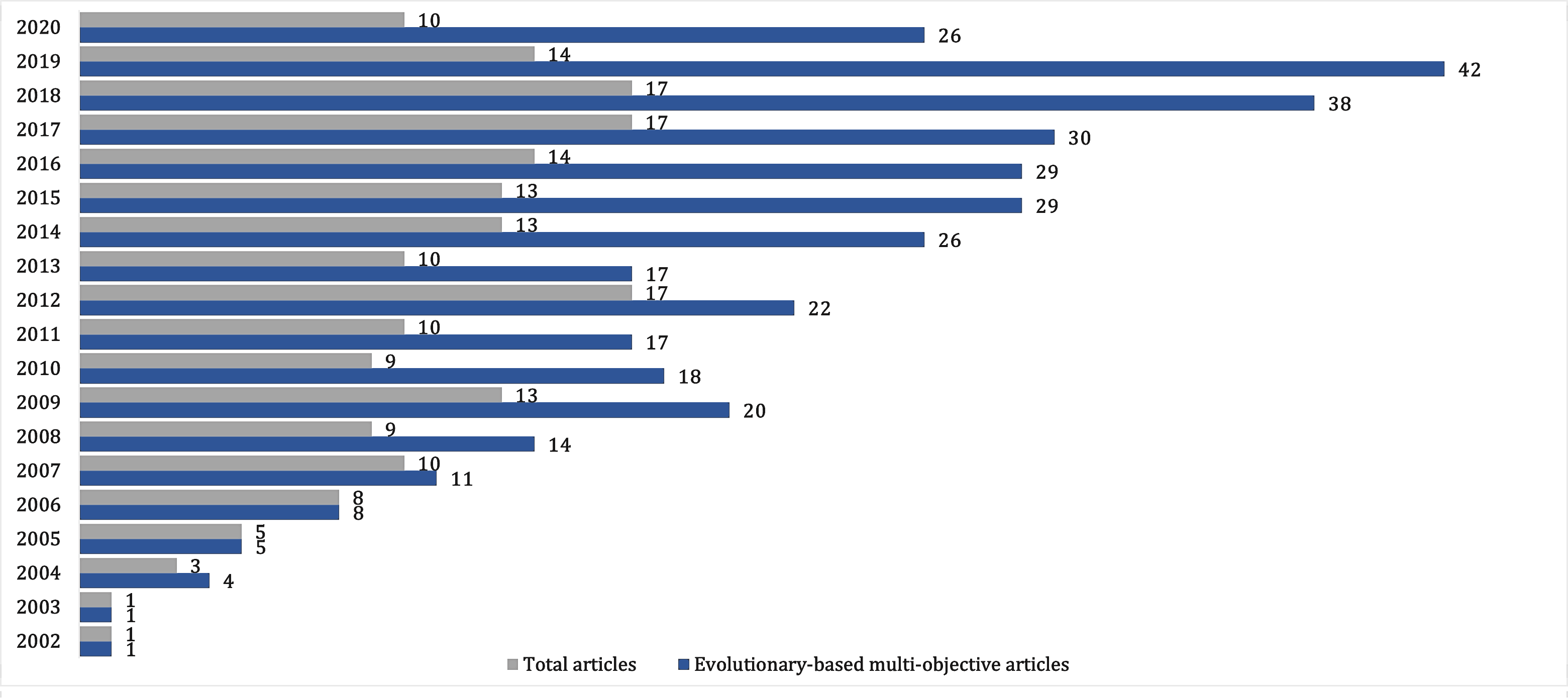}
  \caption{Total articles vs. evolutionary-based optimization articles.}
  \label{fig:articlesGa}
\end{figure}
Regarding the optimization approach, considering the general classification of the metaheuristics presented by \cite{siarry2016metaheuristics}, we observed that most studies applied evolutionary optimization. 
Fig.~\ref{fig:articlesGa} presents the  relationship between the number of articles and the evolutionary optimization articles, including memetic and hybrid approaches that include other methods associated with the evolutionary approach.  
In the early years, almost all MOC papers  relied on the evolutionary approach. In the middle years, the use of other optimization methods was observed, such as Artificial Immune system-inspired~\cite{TIMMIS200811}, Differential Evolution-based~\cite{Eltaeib2018}, Simulated Annealing-based~\cite{bertsimas1993simulated}, and Particle swarm-based~\cite{rana2011review}. In the mapped articles, the first occurrence of these approaches was between 2007 and 2009. In  recent years, the use of a variety of other optimization methods has also been verified, such as other nature-inspired algorithms~\cite{siarry2016metaheuristics}, among others.

\begin{figure*}[!htb]
  \centering
  \includegraphics[width=\textwidth]{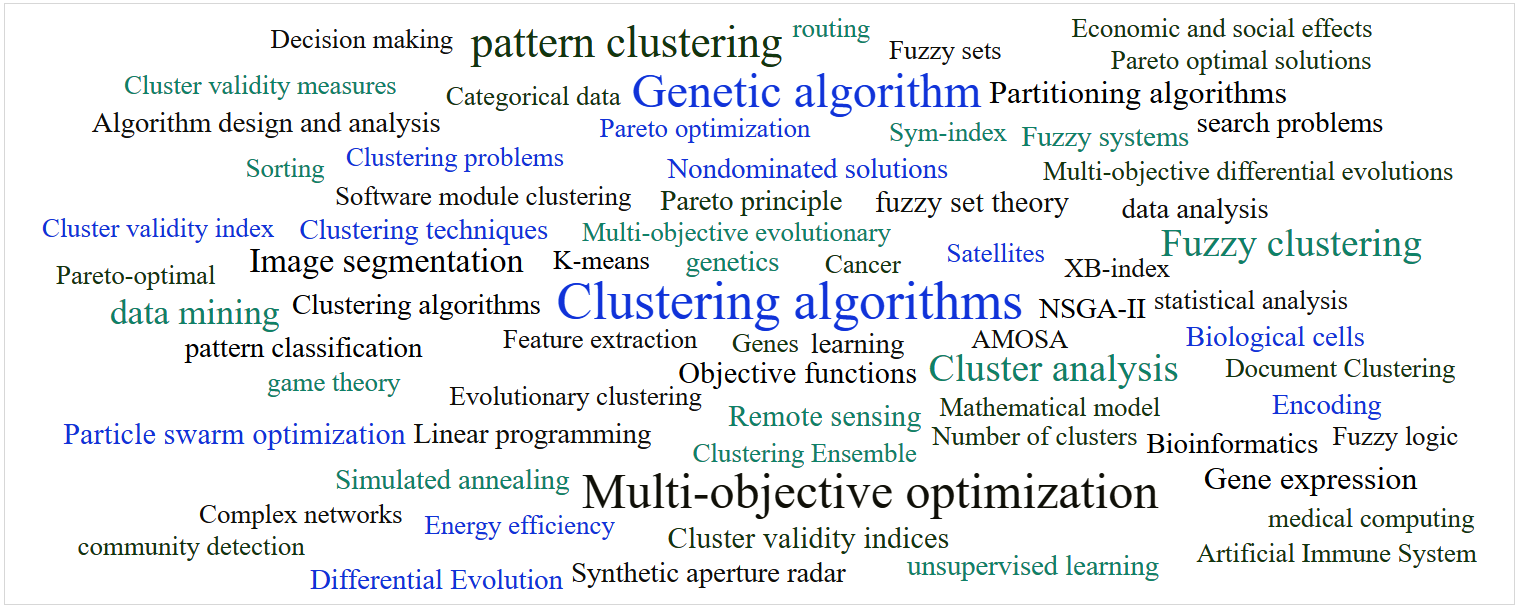}
  \caption{Word cloud of the keywords and the indexed terms in MOC papers}
  \label{fig:wc}
\end{figure*}

We also verified the main topics considered in the almost twenty years of research in MOC. Fig.~\ref{fig:wc}  presents the word cloud of the keywords and the indexed terms in the MOC papers. 
Most of the terms refer to the optimization methods and application  fields of the MOC.  
The same meaning terms (single/plural and case forms of the terms) in the cloud are filtered, so "genetic algorithm", "Genetic Algorithm", and "genetic algorithms" are all treated as "Genetic algorithm". The larger words in the word cloud are the ones used more frequently in the papers analyzed.From this word cloud, although keywords related to the algorithm such as "Clustering algorithms" and "multi-objective optimization" are understandably used more frequently, it is notable that the keywords "multi-objective optimization" and "Genetic algorithm"  have been attracting researchers' attention as a problem domain of MOC. 
We also observe two main application fields: Image Segmentation
and Gene/Micro-Array Analysis. The second one considers words like  "Biological Cell" and "Gene Expression". Other listed application fields are: Document Clustering, Community Detection, Software Module Clustering, among others.

\section{A Literature Review of Evolutionary Multi-objective Clustering Algorithms}\label{sec:MCA}
In this section, we present EMOC algorithms. The most relevant works were selected by considering two general indices: h-index and Scopus-percentile. We  filtered the articles by h-index greater than 10 to filter the conference papers and by Scopus-percentile greater than 50\% to obtain the list of the most relevant journal papers. These values were selected to cover the A-rank papers in the CORE - Computing Research and Education Association of Australasia and Qualis (a Brazilian official system to classify scientific production). 
These algorithms were grouped based on some shared characteristics that highlight the main features or applications of these approaches. The general concepts and methods applied in these EMOC approaches were introduced in Section~\ref{sec:MC}.

\subsection{General-purpose EMOCs algorithms}
First, we present general-purpose EMOC approaches divided in: MOCK-based works, EMOC for categorical data, EMOC for bi-clustering, EMOC for subspace clustering, ensemble-based EMOC, fuzzy clustering-based EMOC, spectral clustering-based EMOC, multiple distance measures-based EMOC, multi-k-clustering-based EMOC, EMOC with  specific MOEA, and other EMOC approaches.

\subsubsection{MOCK-based works} 
One of the most popular algorithms is MOCK  - Multi-Objective Clustering with automatic $k$-determination~\cite{Handl2005b, Handl2005c, Handl2007}. The MOCK algorithm uses LAG representation, initialization with MST-clustering and KM, and two objective functions: $Dev$ and $Con$. The PESA-II was the MOEA  used in this approach. The adjacency graph representation promoted the use of specific operators for the clustering problem, such as the neighborhood-based mutation operator, which  manipulates the links over the MST, in which each vertex can only be linked to one of its nearest neighbors. After the optimization process and the generation of final clustering solutions, MOCK uses an automatic $k$-determination scheme to choose the best clustering solution from a set of solutions with a knee-based strategy. 

Other studies were derived from the analysis of MOCK, as follows. Matake et al.~\cite{Matake2007} provided an approach, MOCK-Scalable, to improve  the final selection of solutions in large-scale data based on a scaling filter to reduce the solutions in the Pareto front. Tsai et al.~\cite{Tsai2012} proposed the MIE-MOCK - Multiple Information Exchange Multi-Objective Clustering with automatic $k$-determination.  The MIE-MOCK algorithm uses a pool of crossover and mutation operators selected by a random method and also provides a new final selection of solutions based on two CVIs: $PMB$ and $DB$.
In~\cite{handl2012clustering}, Handl and Knowles analyzed four pairs of objective functions for multi-objective clustering, including an analysis of the original objective functions of MOCK.
In~\cite{handl2013evidence}, Handl and Knowles analyzed the use of evidence accumulation to support the post-processing of the clustering solutions returned by the MOCK.
In \cite{garza2017, Garza2018}, the authors proposed the $\Delta$-MOCK,  providing a new encoding to improve the MOCK scalability and other specific modifications to improve the convergence of the solutions.
Zhu et al.~\cite{Zhu2018} provided the $\Delta$-EMaOC  - Evolutionary Many-Objective Optimization Clustering, improving the general architecture of the  $\Delta$-MOCK to optimize five objective functions. The $\Delta$-EMaOC algorithm considers the use of MaOEAs (SPEA-II-SDE \cite{Li2014}, NSGA-III~\cite{Yuan2016}, MOEA/DD~\cite{Li2015b} and RVEA~\cite{Cheng2016}) instead of MOEA (NSGA-II). In general, these approaches are applied to detect clusters in heterogeneous structured data, considering a continuous data type and crisp clustering.

\subsubsection{EMOC for Categorical Data}
In particular, some EMOC approaches were designed for categorical data clustering, where the data objects are defined over categorical attributes (instead of using the continuous data type that is applied in most of the other approaches). 
For example, Handl and Knowles~\cite{Handl2005b} presented the MOCK-medoid, a MOCK extension for multi-objective clustering around medoids for categorical data. Mukhopadhyay and Maulik~\cite{Mukhopadhyay2007} also introduced a medoid-based EMOC, the MOGA-medoid, to deal with categorical data. The MOGA-medoid algorithm uses the NSGA-II to optimize the $Sil$ and $Dev$ (computed in terms of the medoid instead of the centroids), 
applying the one-point crossover and a medoid-based replacement mutation designed to consider a center-based solution encoding.

Dutta et al.~\cite{dutta2012data} provided a specific MOEA, the Hybrid MOGA, to optimize $H$ and $Sep_{AL}$. The main contribution of this work relies on the use of this new MOEA with the Pairwise Crossover~\cite{Franti1997}, the replacement (substitution) Mutation, and the local searching power of K-modes (or KM) to deal with continuous and categorical features in the dataset. 

Mukhopadhyay et al.~\cite{Mukhopadhyay2007icit} presented a multi-objective genetic fuzzy clustering of categorical attributes (MOGA-fuzzy), considering a uniform crossover and a center-based replacement mutation in NSGA-II to optimize the global compactness (a normalized $J_m$ index for categorical data)~\cite{tsekouras2004fuzzy} and $Sep_{fuzzy}$. They applied a  specific selection method to obtain the final solution, in which the points assigned to the same cluster by at least 50\% of the clustering solutions are taken as the training set, and the remaining points are assigned a class label using $k$-nearest neighbor ($k$-nn) classification  in order to select a single solution from the set of the non-dominated solutions. In~\cite{Mukhopadhyay2009}, the authors provide a new version of the MOGA-fuzzy (MOGA-fuzzy2), considering modifications in the evolutionary operators, in which the One-Point Crossover and Mode replacement were applied. 

Zhu and Xu~\cite{Zhu2018_MaOEA} introduced the MaOFcentroids, a many-objective fuzzy centroid clustering algorithm for categorical data. MaOFcentroids algorithm uses fuzzy membership matrix encoding (a matrix with the degree of membership of each object), and adapted operators that consider the number of the clusters and the membership of the solutions in the NSGA-III. It simultaneously optimizes five CVIs ($CDCS$, $DB$, $CH$, $CCI$, and $XB$). In terms of the selection, this approach uses a specific clustering ensemble for categorical data, the SIVID - Sum of Internal Validity Indices with Diversity~\cite{Zhao2017}.

The most recent work of Dutta et al.~\cite{dutta2019automatic} introduces the MOGA-KP, an approach with automatic $k$-determination applied to deal with different types of features (continuous, categorical, and missing feature values). It considers some common aspects of the previous works~\cite{dutta2012data, dutta2012simultaneous}, while improving some aspects, such as the use of other evolutionary operators, and the local search operators. Besides that, the MOGA-KP algorithm uses a ranking of nine CVIs ($C$ index, $COSEC$, $DB$, $Dunn$, $Dev$, $Entropy$, $XB$, $Purity$, $F$-Measure) to determine the final set of solutions. 

\subsubsection{EMOC for Bi-Clustering}
One specific line of  study in EMOC is Bi-clustering, which consists of simultaneous partitioning of the set of samples and the set of their attributes into subsets (classes). The goal of this kind of algorithm is to find one or all (possibly overlapping) sub-matrices of a given matrix, each of which shares a pre-defined property over the elements across all its columns (or rows). Each such sub-matrix is called a bi-cluster.
Bousselmi et al.~\cite{bousselmi2017bi} presented the BI-MOCK, which extends MOCK to the case of bi-clustering by adding a subset of columns (conditions) to each chromosome in the representation. BI-MOCK algorithm uses the  two-points crossover adapted for variable-size chromosomes and the CC algorithm as a mutation operator in the PESA-II to optimize  $Var$, and the size of the bi-cluster.
Bechikh et al.~\cite{bechikh2019hybrid} presented the MOBICK - Multi-Objective BI-Clustering with automated $k$ deduction, that extends Bousselmi et al.~\cite{bousselmi2017bi} study. MOBICK algorithm uses the $\Delta$-MOCK reduced encoding, the uniform crossover adapted for bi-clustering conditions, and the CC algorithm as the mutation operator in the PESA-II to also optimize $Var$ and the size of the bi-cluster. 

\subsubsection{EMOC for Subspace Clustering}
Another line of studies considers Subspace Clustering, an extension of traditional clustering that seeks to find clusters in different subspaces within a dataset.
Zhu et al.~\cite{Zhu2012} introduced the MOSSC - Multi-Objective evolutionary algorithm-based Soft Subspace Clustering, which optimizes the $SSBX$ and $J_{wm}$ in the NSGA-II. This approach uses a center-based encoding with weights to avoid trapping in local minima, aiming to obtain more stable clustering results. 
Xia et al.~\cite{xia2013novel} presented the MOEASSC - Multi-Objective Evolutionary Approach-based Soft Subspace Clustering, which also uses a mixed encoding (center and weight-based). MOEASSC differs from the MOSSC in terms of the pair of objectives ($J_{wm}$ and $J_{Add}$), and the use of a local search operator based on the KM. 
Zhou and Zhu \cite{zhou2018kernel} introduced the MOKCW - Multi-Objective Kernel Clustering algorithm with automatic attribute Weighting. In general, MOKCW extends MOSSC and MOEASSC by considering kernel clustering. For example, MOKCW used the MOSSC objective functions adapted to consider kernel distance. The authors also improved the final selection method of the MOEASSC by applying a clustering ensemble method (MCLA and HBGF) associated with the PSVIndex.

\subsubsection{Ensemble-based EMOC}
Another specific approach was proposed by Faceli et al.~\cite{Faceli2006}, the MOCLE - Multi-Objective Clustering Ensemble. The main idea  behind this approach is the use of clustering ensemble methods as crossover operators to combine partitions and extract agreed patterns to generate new solutions in the evolutionary optimization process. MOCLE is a framework that uses a label-based representation; the initial population is generated with various clustering methods to detect different cluster formats, such as SL, AL, KM, and SNN. 
The original implementation of the MOCLE~\cite{Faceli2006} provides two MOEAs: NSGA-II and SPEA-II, to optimize the $Dev$ and $Con$; and two crossover operators: MCLA - Meta Clustering Algorithm~\cite{Strehl2002} and HBGF - Hybrid Bipartite Graph Formulation~\cite{Fern2004}; however it does not use any mutation operator. 

This general concept of using clustering ensemble methods as crossover operators has been used in other  studies as well. Faceli et al.~\cite{Faceli2009} introduced the MOCLE in the context of gene expression datasets, applying an additional objective, $ConP$ (the connectivity index based on the Pearson Correlation)  and a new set of clustering methods to generate the initial population (AL, CL, KM, and SPC). Liu et al.~\cite{Liu2012} introduced the IMOCLE - Improvement of the Multi-Objective Clustering Ensemble algorithm, in which a relative CVI, $Sim$, was added along with the three objective functions defined by Faceli et al.~\cite{Faceli2009} to improve the clustering. In general, these approaches are also applied to detect clusters in heterogeneous structured data, considering continuous data type and crisp clustering.

\subsubsection{Fuzzy Clustering-based EMOC}
Another line of studies considers the integration of the general concepts of the existing fuzzy clustering algorithms,  such as FCM and FRC - Fuzzy Relational Clustering, with a multi-objective evolutionary approach (NSGA-II).
Di Nuovo et al.~\cite{DiNuovo2007},  Wikaisuksakul~\cite{Wikaisuksakul2014} and Dong et al.~\cite{dong2018adaptive} presented fuzzy approaches integrating the NSGA-II with the FCM~\cite{bezdek2013pattern}.
Di Nuovo et al.\cite{DiNuovo2007} introduced the NSGA-II\&FCM that optimize the number of features and the $XB$ index to discover the best number of groups while pruning the features to reduce the dimensionality of the dataset. NSGA-II\&FCM algorithm uses a specific solution encoding that considers the FCM parameters (number of the cluster $k$ and FCM fuzzyfier $m$) and the feature weights.
Wikaisuksakul~\cite{Wikaisuksakul2014} introduced the FCM-NSGA, which optimizes the $J_m$ and $Sep_{nfuzzy}$ in NSGA-II, considering SBX and polynomial mutation operators.
Dong et al.~\cite{dong2018adaptive} introduced the ADNSGA2-FCM that optimizes the $DB$ and $\mathcal{I}$ indexes. ADNSGA2-FCM uses a center-based and fuzzy membership matrix (a matrix with the degree of membership of each object) as an encoding. In terms of the evolutionary operators, it considers the uniform mutation with two new crossover operators, the Nearest Neighbor Matching Crossover Operation (an exchange of centers in the nearest neighbor to produce solutions with the same number of clusters) and  the Truncation and Stitching Crossover Operation (an exchange of a set of center positions is performed to produce solutions with a different number of clusters). Moreover, they introduced an adaptive  mechanism that is applied to compute the crossover and mutation probabilities that are changed according to the fitness of the current population.
On the other hand, Paul and Shill~\cite{paul2018new} propose the FRC-NSGA/IFRC-NSGA, hybrid methods that combine the FRC algorithm~\cite{Skabar2013} and the NSGA-II to optimize the $J_m$ and $Sep_{nfuzzy}$.

\subsubsection{Spectral Clustering-based EMOC}
Some works use spectral clustering as a foundation for designing EMOC approaches. 
MOGGC - Multi-Objective Genetic Graph-based Clustering Algorithm~\cite{menendez2013multi}
considers optimizing the computation of graph similarity features in SPC to achieve lower memory consumption and increase the clustering quality. For that, this approach provided a new objective function pair,  the separation of clusters ($Sep_{Graph}$) and a graph continuity metric (DCD). MOGGC was extended by the CEMOG - CoEvolutionary Multi-Objective Genetic Graph-based Clustering~\cite{menendez2014co}, a partitional $k$-adaptive spectral clustering algorithm that uses a strategy based on island-model and a graph topology to migrate individuals from sub-populations. This last approach does not require input of the initial number of clusters required in the MOGGC.
In this context, Luo et al.~\cite{Luo2015} introduced the framework SRMOSC, which uses sparse representation for sparse spectral clustering. SRMOSC uses the $SP$ and $RE$ as objective functions to be optimized in the NSGA-II (or MOEA/DD) with a specific pair of operators that considers the sparsity properties. 

\subsubsection{Multiple Distance Measures-based EMOC}
Other approaches consider the use of different distance functions in the objective functions.
Liu et al.~\cite{Liu2018} introduced the MOECDM - Multi-objective Evolutionary Clustering Based on Combining Multiple Distance Measures and the MOEACDM - Multi-objective Evolutionary Automatic Clustering based on Combining Multiple Distance Measures. Both these approaches consider a single CVI computed with distinct distance functions to define objective functions to be optimized. They use a label-based encoding and an NCUT pre-clustering\cite{shi2000} in the initialization, but in the MOECDM, a portion of the individuals are generated by a random generator. They also adapted the crossover and mutation operators, in which the probabilities are adjusted along with the generations. MOECDM was designed to detect the desirable cluster number automatically, using the $Sep_{CL}$ index computed with Euclidean distance ($Sep_{CL1}$) and Path distance ($Sep_{CL2}$) as objective functions. MOEACDM was designed to detect compact clusters, using the $Mod$ also computed with Euclidean distance ($Mod1$) and Path distance ($Mod2$) as objective functions. 

\subsubsection{Multi-k-clustering-based EMOC}
Other approaches consider multi-k-clustering with the a posteriori method, where $k$ is taken as an objective function, differing from the automatic data clustering methods, such as MOCK, that consider $k$ an inner aspect of the decision variable, obtained by the optimization of clustering criteria.
For that, Du et al.\cite{Du2005} introduced a specific solution representation, the linked-list based encoding.  The authors used the fellowship between the objects instead of the label-based relationship to define the clusters, in which each cluster has all  its  elements linked, similar to the  relationship  of the nodes presented by Handl and Knowles~\cite{Handl2005b}. This representation was applied in the MOGA-LL\cite{Du2005}, an EMOC approach that optimizes the $TWCV$ and $k$ as objective functions in the NPGA, considering two particular operators: (i) an adapted one-point crossover, which allows different clusters to exchange partial contents and may split a cluster into two; (ii) link-replacement mutation, in which a sub-group of objects is associated  with another cluster instead of just a different node.  

Wang et al.~\cite{wang2018} proposed the EMO-KC (Evolutionary Multi-objective $k$-clustering) to demonstrate the importance of the conflict between the objective functions to obtain a diverse set of final solutions with a different number of clusters. They showed evidence that $Var$ and $k$ are not always conflicting between two individuals and introduced a transformation of the variance ($Var$) formulation, $(1- exp^{-1.Var})-k$, to solve this problem. In ~\cite{WANG2020100691}, this same pair of objective functions was explored in a new MOEA that considers a constrained decomposition with grids (CCDG-K). Both EMO-KC and CCDG-K define the best clustering result (the optimal $k$) by the ``elbow'' method~\cite{HANCER201749}.

\subsubsection{Specific MOEA for EMOC}
 As previously presented, Dutta et al.~\cite{dutta2012data, dutta2012simultaneous} provided a specific MOEA, the Hybrid MOGA designed for categorical data. Besides that, another particular approach is the VRJGGA - Variable-length Real Jumping Genes Genetic Algorithm introduced by Ripon et al.~ \cite{Ripon2006_ICPR}. The VRJGGA is an EMOC algorithm that extends  the  Jumping Genes Genetic Algorithm  (JGGA)~\cite{Man2004_jump} and  applies  the survival selection of the NSGA-II. 
The JGGA considers jumping gene operations before evolutionary operators to improve the diversity of solutions. 
VRJGGA uses a centroid-based encoding associated with the modulo crossover~\cite{Srikanth1995} (an adapted one-point crossover, where each child is a set of  completely specified sub-solutions) and the polynomial mutation, to optimize the $Ent$ and $Sep_AL$.
In~\cite{Ripon2006_IJCNN}, the authors provided new features to VRJGGA, introducing two local search methods, probabilistic cluster merging, and splitting for clustering improvement.
Ripon and Siddique~\cite{Ripon2009} also applied the extended version of the JGGA to EMOC, introducing the EMCOC - Evolutionary Multi-objective Clustering for detecting overlapping clusters.  EMCOC introduces a new chromosome representation and cluster-assignment method in which each data point is a candidate center and a binary encoding is applied to define whether a data point is a center or not. 

\subsubsection{Other MOC approaches}
Some works consider other objective functions and provide other features in the design of the EMOC approaches. For example, Kirkland et al.~\cite{Kirkland2011} presented the Multi-Objective Clustering algorithm (MOCA), that optimizes three objective functions, $AWGSS$, $ABGSS$, and $Con$ in the NSGA-II.
Sert et al.~\cite{Sert2011, sert2012unification} presented the MOC-HCM,  which uses five objective functions:  Km\textsubscript{id}, Km\textsubscript{ed}, Km\textsubscript{wid}), Km\textsubscript{wed}, and EWCD. The MOC-HCM algorithm uses a binary representation,  a local search operator (k-mode-based operator) that reassigns the instances to the closest clusters in terms of their frequencies, and a new final selection method based on a new metric, the H-Confidence Metric (HCM). 

Besides the above-mentioned works, we also found specific approaches, in which their main features consider some particular methods, as follows.  {\"O}zyer and Alhaj~\cite{ozyer2009parallel} applied the divide and conquer approach in an iterative way to handle the clustering process and improve the performance  of the evolutionary algorithm. Zheng et al.~\cite{zheng2012multi} extended algebraic operations of gene expression to propose a multi-objective gene expression programming for clustering.  Garcia-Piquer~\cite{garcia2017scaling}, focused on reducing the impact of the volume of data in the EA by means of the stratification of the complete data set into disjoint strata and  alternating them in each cycle of the GA. Liu et al.~\cite{liu2019transfer} improved the performance of multi-objective soft subspace clustering algorithms for clustering high-dimensional data by  using  a transfer learning-assisted multi-objective evolutionary clustering framework with MOEA/D.

\subsubsection{Summary of the EMOC approaches}

Here, we summarize the components of the main presented EMOC algorithm.
We considered the publishing chronology to list each EMOC to make it possible to observe the variations of components over time.

In Table \ref{tab:comparacao}, we present the main features (components) applied in the initialization and optimization of each approach. In this table, we used acronyms and abbreviations for some words: Ad. for Adapted, Repl. for Replacement, and Mod. for Modified, NA for not assigned, and FM for Fuzzy membership-based. 

\afterpage{
\begin{landscape}
{\footnotesize
\centering
\begin{longtable}{p{.03\columnwidth}p{.13\columnwidth}>{\raggedright\arraybackslash}p{.11\columnwidth}p{.105\columnwidth}p{.10\columnwidth}>{\raggedright\arraybackslash}p{.16\columnwidth}p{.215\columnwidth}}
\caption*{EMOC algorithms: initialization and optimization components}\label{tab:comparacao} \\\hline
    \multicolumn{1}{l}{\textbf{Year}} &
    \multicolumn{1}{l}{\textbf{Article}} &
    \multicolumn{1}{l}{\textbf{Representation}} &
    \multicolumn{1}{l}{\textbf{Initialization}} & 
    \multicolumn{1}{l}{\textbf{MOEA}} & 
    \multicolumn{1}{l}{\textbf{Objectives}}& 
    \multicolumn{1}{l}{\parbox{3.5cm}{\textbf{Crossover/Mutation}}} \\\hline
\endfirsthead
            \multicolumn{6}{c}
            {{ \tablename\ \thetable{} -- continued from previous page}} \\ \hline 
				   
             \multicolumn{1}{l}{\textbf{Year}} &
            \multicolumn{1}{l}{\textbf{Articles}} & 
            \multicolumn{1}{l}{\textbf{Representation}} & \multicolumn{1}{l}{\textbf{Initialization}} & 
            \multicolumn{1}{l}{\textbf{MOEA}} & 
            \multicolumn{1}{l}{\textbf{Objectives}} & 
            \multicolumn{1}{l}{\parbox{3.5cm}{\textbf{Crossover/Mutation}}} \\ \hline
            \endhead
            \hline
            \endfoot
\endlastfoot         
    2005 & MOCK \cite{Handl2005b, Handl2005c, Handl2007} & LAG & MST, KM & PESA-II & ($Con$, $Dev$)   & Uniform/Neighborhood-based  \\
    2005 & MOCK-medoid \cite{Handl2005b} & LAG &  KM & PESA-II & ($Var$, $Dev$) & Uniform/Neighborhood-based \\
    2006 & VRJGGA \cite{Ripon2006_ICPR, Ripon2006_IJCNN} & Centroid-based & Random & JGGA-based  & ($Ent$, $Sep\textsubscript{AL}$)  & Modulo/Polynomial  \\
    2006 & MOCLE \cite{Faceli2006} & Label-based & KM, AL, SL, SNN & SPEA2/NSGA-II & $(Con$, $Dev$) & HBGF or MCLA/---  \\
	2007 & MOGA-medoid~\cite{Mukhopadhyay2007} & Medoid-based & Random & NSGA-II & ($Dev$, $Sil$)  & One-point/Medoid Repl. 	\\
    2007 & MOCK-scalable \cite{Matake2007} & LAG & MST , KM & SPEA2 &  ($Con$, $Dev$) & Uniform/Neighborhood-based  \\
	2007 & MOGA-fuzzy \cite{Mukhopadhyay2007icit} & Mode-based & Random & NSGA-II & ($J_m$, $Sep\textsubscript{fuzzy}$) & Uniform/Mode Repl. \\
	2007 & NSGA-II\&FCM \cite{DiNuovo2007} &FCM parameters and features weights & FCM & NGSA-II & (Number of features, $XB$) & SBX/Polynomial \\
	2009 & MOGA-fuzzy2 \cite{Mukhopadhyay2009} &  Mode-based & Random & NSGA-II & ($J_m$, $Sep_{fuzzy}$) & One-Point/Mode Repl.  \\
	2009 & EMCOC \cite{Ripon2009} & Binary center-based &Random & JGGA-based  & ($Ent$, $Sep\textsubscript{AL}$)  &  Center exchange /NA  \\
	2011 & MOCA \cite{Kirkland2011} & Medoid-based  & Random & NSGA-II & ($AWGSS$, $ABGSS$, $Con$) & Centroids exchange/ Split, Merge, Centroid Repl. \\
	2011 & MOC-HCM \cite{Sert2011, sert2012unification} & Binary-based & NA & NSGA & ($Km\textsubscript{id}$, $Km\textsubscript{ed}$, $Km\textsubscript{wid}$, $Km\textsubscript{wed}$, $EWCD$) & Shuffle/NA  \\
    2012 & IMOCLE \cite{Liu2012} & Label-basaed & KM, AL, CL, SPC & NSGA-II & ($Con$, $ConP$, $Dev$, $Sim$) & MCLA/---   \\
    2012 & Hybrid MOGA \cite{dutta2012data, dutta2012simultaneous} & Centroid-based & Random & specific MOEA & ($Sep_{CL}$, $H$) & Pairwise/Centroid Repl.     \\    
   2012 & MOSSC \cite{Zhu2012}  & Center and weight-based& Random & NSGA-II & ($SSXB$, $J_{wm}$) & SBX/Polynomial \\
    2012 & MIE-MOCK \cite{Tsai2012} & LAG &MST , KM &PESA-II & ($Con$, $Dev$) & Uniform, One-Point and Two-Point/ Neighborhood-based, Split and Merge \\
	2013 & MOGGC \cite{menendez2013multi} & Label-based & Random & SPEA2 & ($sep_{graph}$, $DCD$) & Labels Exchange/Adaptive \\
	2013 & MOEASSC \cite{xia2013novel} & Center and weight-based &Random  & NSGA-II & ($J_{wm}$, $J_{Add}$)  &One Point/Uniform  \\
	2014 & CEMOG \cite{menendez2014co} & Label-based & Random & SPEA2 & ($sep_{graph}$, $DCD$ & Labels Exchange/Adaptive  \\ 
	2014 & FCM-NSGA \cite{Wikaisuksakul2014}& Center-based &Random & NSGA-II & ($J_{m}$, $Sep_{nfuzzy}$) & SBX/Polynomial	\\ 
	2016 &SRMOSC \cite{Luo2015}& Sparse-based &  Neighbor-based&  NSGA-II/ MOEA/DD& ($RE$, $SP$) & specific operators based on the sparsity property  \\ 
    2017 & $\Delta$-MOCK \cite{garza2017, Garza2018}& Reduced LAG & MST & NSGA-II & ($Con$, $Var$) & Uniform/Neighborhood-based  \\
    2017& BI-MOCK \cite{bousselmi2017bi}& LAG with conditions & MST, KM & PESA-II &  ($Var$, size of the bi-cluster) & Ad. Two-Points/ CC  \\
    2018 & MOKCW \cite{zhou2018kernel} & Center and weight-based & Random &  NSGA-II & (Ad. $J_{wm}$, Ad. $SSBX$) & One-Point/Uniform\\
	2018 & EMO-KC \cite{wang2018} & Centroid-based&Random &NSGA-II & (Mod. $Var$, $k$) & SBX/ Polynomial \\
	2018& FRC-NSGA \cite{paul2018new} & Center-based & Random & NSGA-II & $J_m$ and $Sep_{nfuzzy}$& SBX/Polynomial  	\\
    2018& $\Delta$-EMaOC \cite{Zhu2018} & Reduced LAG &MST & NSGA-III/ RVEA/ MOEA/DD/ SPEA-II-SDE  & ($Con$, $Var$, $Dunn$, $DB$, $CH$) & Uniform/Neighborhood-based\\
	2018& MOECDM \cite{Liu2018} & Label-based &Random, NCUT & NSGA-II & ($Sep_{CL1}$, $Sep_{CL2}$)  & Ad. Uniform/Ad. Uniform  \\
	2018&  MOEACDM \cite{Liu2018} & Label-based &NCUT & NSGA-II & ($Mod1$, $Mod2$)  & Ad. Uniform/ Ad. Uniform  \\
	2018& ADNSGA2-FCM \cite{dong2018adaptive} & Center and FM-based & Random & NSGA-II & $DB$ and $\mathcal{I}$ & Neighborhood-based, Truncation and Stitching/ Uniform  \\
    2018& MaOFcentroids \cite{Zhu2018_MaOEA}&FM-based& Random & NSGA-III & ($CDCS$, $DB$, $CH$, $CCI$, $XB$) &  Uniform/specific Mutation for Membership Repl.  \\
	2019& MOBICK~\cite{bechikh2019hybrid}& Reduced LAG with conditions& MST, KM & PESA-II &  ($Var$, size of the bi-cluster) & Ad. Uniform/CC \\
	2019& MOGA-KP~\cite{dutta2019automatic}& Centroid-based & Random & specific MOEA& ($Sep_{CL}$, $H$)  &One-Point/Polynomial \\
\hline
\end{longtable}}
\end{landscape}

}

It is possible to note that there are a variety of representations being applied in the EMOC approaches. In particular, from the year 2017, the use of representations concerning the reduction of the size of the chromosome has emerged. In contrast, most EMOC approaches use a random strategy in the initialization, without introducing a relevant novelty in recent years.

Regarding the optimization phase, the NGSA-II has been the most applied MOEA over the years. In particular, from the year 2018, the use of MaOEAs considering the optimization of more than 3 objective functions has emerged. 
In terms of the objective functions, over the years, new combinations of clustering criteria have been applied. A common practice considers at least one compactness-based criterion associated with a connectedness-based criterion for clustering heterogeneous structured data.  In the case of the centered-based clustering optimization, it is common to see other schemes for the objectives:  (i) a compactness-based criterion and the number of the clusters, (ii) a combination of the two compactness-based criteria,  (iii) a compactness-based criterion and a spatial separation-based criterion. In this last case, these different configurations of objective functions are mostly related to specific classes of clustering studies, such as bi-clustering (i), categorical data clustering (ii and iii).
The same  occurs with the crossover and mutation operators, in which we can observe a diversity of combinations of operators.

Table \ref{tab:comparacao3}  summarizes  the selection methods applied to each approach that provides this component in their design. As this component is not mandatory in the EMOC design, almost half of the presented algorithms do not provide it. The existing selection methods are, in general, as follows: ensemble-based, which provides the best solution (consensual partition); knee-based, which provides the best k-solution; and CVIs-based, which  considers specific criteria (as ranking) to define the best set of solutions.

\begin{table}[!ht]
\centering
\caption{EMOC algorithms: selection strategies}\label{tab:comparacao3} 
\footnotesize
\begin{tabular}{p{.05\columnwidth}p{.325\columnwidth}>{\raggedright\arraybackslash}p{.455\columnwidth}}\hline
    \multicolumn{1}{l}{\textbf{Year}} &
    \multicolumn{1}{l}{\textbf{Article}} &
    \multicolumn{1}{l}{\textbf{Final Selection}} \\\hline
    2005 & MOCK \cite{Handl2005b, Handl2005c, Handl2007} & Knee-based \\
    2005 & MOCK-medoid \cite{Handl2005b} & Knee-based\\
    2007 & MOCK-scalable \cite{Matake2007} &  Knee-based  \\
	2007 & MOGA-fuzzy \cite{Mukhopadhyay2007icit} &  Specific approach ($k$-nn-based)  \\
	2009 & MOGA-fuzzy2 \cite{Mukhopadhyay2009} & Specific approach ($k$-nn-based)  \\
	2011 & MOC-HCM \cite{Sert2011, sert2012unification} & Ensemble-based (H-confidence) \\
    2012 & MIE-MOCK \cite{Tsai2012} &   PBM and DB \\
    2012 & MOSSC \cite{Zhu2012}  & Ensemble-based (HBGF)\\
	2013 & MOGGC \cite{menendez2013multi} & $sep_{graph}$ \\
	2013 & MOEASSC \cite{xia2013novel} & PSVIndex \\
	2014 & CEMOG \cite{menendez2014co} &  $sep_{graph}$ \\ 
	2016 &SRMOSC \cite{Luo2015}&  Ratio cut-based\\ 
    2018 & MOKCW \cite{zhou2018kernel} &  PSVIndex and ensemble-based (HBGF or MCLA)\\
	2018 & EMO-KC \cite{wang2018} & Elbow-based \\
	2018& ADNSGA2-FCM \cite{dong2018adaptive} &  Ensemble-based  (Majority vote) \\
    2018& MaOFcentroids \cite{Zhu2018_MaOEA}&  Ensemble-based (SIVID) \\
	2019& MOGA-KP~\cite{dutta2019automatic}&   $DB$, $Dev$, $Dunn$, $C$, $COSEC$, $Entropy$, $F$-Measure, $Purity$ and $XB$\\
\hline
\end{tabular}
\end{table}

\subsection{EMOC Approaches Designed for Specific Applications}\label{sec:Application}
In this section, we present approaches designed for specific applications. Each algorithm considers the particularities of problem application to define the representation of the solutions, the objective 
functions, or/and the evolutionary operators. It promotes the generation of a variety of configurations, so we will limit ourselves to listing some algorithms designed for each following application.

\subsubsection{Association rule learning} Association rule learning is a rule-based machine learning method for discovering interesting relations between variables in large databases. Alhajj and Kaya  ~\cite{Kaya2004, alhajj2008multi} provided an EMOC approach for fuzzy association rules mining to automatically cluster values of a given quantitative attribute to obtain a high number of large itemsets in low duration (time). 

\subsubsection{Document clustering} Document clustering is a data/text mining technique that makes use of text clustering to divide documents according to various topics. Lee et al.~\cite{lee2014document}  proposed a method of enhancing multi-objective genetic algorithms for document clustering with parallel programming. Wahid et al.~\cite{wahid2015multi}  presented a new approach for document clustering based on SPEA-II,  that explores the concept of multiple views to generate multiple clustering solutions with diversity.

\subsubsection{Gene/micro-array analysis} The Gene/Micro-array clustering analysis is applied to discover groups of correlated genes potentially co-regulated or associated with the disease or conditions under investigation. 
Romero-Zaliz et al.~\cite{romero2008multiobjective} provided an EMOC to identify conceptual models in structured datasets that can explain and predict phenotypes in the immune inflammatory response problem, similar to those provided by gene expression or other genetic markers. 
Li et al.~\cite{li2017ensemble} provided a new ensemble operator to improve the data clustering in gene expression datasets in IMOCLE \cite{Liu2012}. Mukhopadhyay et al.~\cite{mukhopadhyay2010simultaneous} provide an approach that simultaneously selects relevant genes and clusters the input dataset. Mukhopadhyay et al.~\cite{Mukhopadhyay2013} presented an interactive approach to multi-objective clustering  of gene expression patterns considering an adapted NSGA-II, in which inputs from the human decision-maker (DM) are taken to learn which objective functions are more suitable for the datasets. 
Dutta and Saha~\cite{dutta2017fusion} presented an EMOC approach to identify gene clusters from a given expression dataset; in which apart from utilizing the gene expression values of the individual genes, the corresponding protein-protein interaction scores are also used while clustering the set of genes.

\subsubsection{Image Segmentation} Image segmentation consists of the process by which a digital image is partitioned into various subgroups (multiple parts or regions), often based on the characteristics of the pixels in the image. Qian et al.~\cite{qian2008unsupervised} presented a multi-objective evolutionary ensemble algorithm to perform texture image segmentation. Shirakawa and Nagao~\cite{shirakawa2009evolutionary} introduced a variation of the MOCK~\cite{Handl2007} improving its general features for its application in image segmentation.
Zhang et al.~\cite{zhang2016multi} provided a multi-objective evolutionary fuzzy clustering for image segmentation, considering the original FCM energy function to preserve image details and a function based on local information to restrain noise, both minimized by MOEA/D. Zhao et al.~\cite{zhao2018noise, zhao2019multi} introduced the use of the concepts of intuitionistic fuzzy set (IFS) and multiple spatial information to generate an EMOC approach to overcome the effect of noise in image segmentation.

\subsubsection{Software module clustering} Software module clustering refers to the problem of automatically organizing software units into modules to improve program structure. Praditwong et al.~\cite{praditwong2010software} provided a multi-objective formulation of the software module clustering problem considering a two-archive Pareto optimal genetic algorithm. Barros~\cite{barros2012analysis} provided an analysis of the effects of composite objectives in multi-objective software module clustering. 

\subsubsection{Network community detection} Network community detection refers to the procedure of identifying groups of interacting vertices in a network depending upon their structural properties to unveil the dynamic behaviors of networks. Folino and Pizzuti~\cite{folino2010multiobjective} provided an approach for the detection of communities with temporal smoothness formulated as an EMOC. Hariz et al.~\cite{hariz2016improving} reformulate the community detection problem as an EMOC model that can simultaneously capture the intra and inter-community structures based on functions inspired by different types of node neighborhood relations. Shang et al.~\cite{shang2017multi} introduced an EMOC approach based on $k$-nodes update policy and a similarity matrix for mining communities in social networks. Pizzuti and Socievole~\cite{pizzuti2019multiobjective} provided a framework for detecting community structure in attributed networks, introducing a post-processing local search procedure that identifies those communities that can be merged to provide higher quality community divisions.

\subsubsection{Web recommendation} Web topic mining and web recommendation consider the problem of extracting web navigation patterns, based on the interests of a user, to be applied in the recommender systems to guide users during their visit to a Web site. Demir et al.~\cite{demir2010multiobjective} presented EMOC approaches to  clustering Web user sessions in a Web page recommender system.
Morik et al.~\cite{morik2012multi} investigated the problem of finding alternative high-quality structures for (Web) navigation in a large collection of high-dimensional data, and they provided a formulation of FTS (Frequent Terms Set) clustering as a multi-objective optimization problem. 

\subsubsection{WSN - Wireless Sensor Network topology management} There are several challenges in designing WSN because the sensor nodes have limited resources of energy, processing power, and memory. In this context, the clustering technique can organize nodes into a set of groups based on a set of pre-defined criteria to improve their usage.  Peiravi et.al~\cite{peiravi2013optimal} provided an EMOC approach whose goal was to obtain clustering schemes in which the network lifetime was optimized for different delay values. 
Hacioglu et al.~\cite{hacioglu2016multi} presented an EMOC approach that can extend network lifetime while enabling high coverage and data. 

\subsubsection{ Other applications}
Wang et al.~\cite{wang2015two} proposed an approach to solve the circuit clustering problem in field-programmable gate array computer-aided design flow.
Bandyopadhyay et al.~\cite{mukhopadhyay2009unsupervisedSVM} introduced a multi-objective genetic clustering  approach for pixel classification in remote sensing imagery.
Wang et al.~\cite{wang2014multiobjective} and Li et al.~\cite{li2016multiobjective} provided a multi-objective fuzzy clustering approach for change detection in Synthetic Aperture Radar (SAR) images. 
Liu et al.~\cite{liu2017shape} presented an approach to automatic clustering of shapes considering a multi-objective optimization with decomposition and improvement in the shape descriptor and diffusion process (that was applied to transform the similarity distance matrix among total shapes of a dataset into a weighted graph).

\section{Conclusion }\label{sec:remarks}

In this paper, we presented a review of the EMOC studies, focused on a general architecture of evolutionary multi-objective clustering, considering the chromosome representation, initialization strategies, MOEAs (or MaOEAs), objective functions, evolutionary operators (crossover and mutation), and final solution selection.  We detailed each feature introducing the main components in the design of EMOC approaches to support the development of studies in this field.   

In particular, this article presented an overview of the publications on multi-objective clustering, considering the indexed papers in the ACM Digital Library, IEEE Xplore, and Scopus.  We selected the papers cited in this review based on metrics of the impact and relevance of the conference/journal, promoting a not-biased selection of papers and better coverage of the EMOC studies. This mapping of EMOC approaches allows us to observe some patterns and obtain some insights regarding the evolutionary multi-objective clustering algorithms. 
For example, the choice of the objective functions is one of the most critical factors in the optimization process. In general, there is no consensus around the ideal number and the best combination of objective functions among researchers because of the difficulty in defining appropriate clustering criteria. In this context, Wang et al.~\cite{wang2018} highlighted the conflict required between the objective functions to generate a diverse and convergence set of solutions, in which the authors provided a modification of $Var$ to improve the conflicting relationship between $Var$ and $k$. On the other hand, MOCK-medoid \cite{Handl2005b} uses similar objective functions, $Var$ and $Dev$, both considering compactness criteria.
In this way, more studies on the objective functions are required to improve the composition of objective functions and provide more information on the limitations of the existing ones.

In terms of an evolutionary multi-objective approach, we can note the wide use of the NSGA-II as MOEAs over the years. In recent years, the use of MaOEAs  has been verified~\cite{Zhu2018, Zhu2018_MaOEA}, in contrast to other works \cite{Sert2011, sert2012unification, Liu2012}  that  considered the optimization of more than three objective functions in MOEAs (NSGA/NSGA-II). However, there are limited studies that analyze the behavior of the other MOEAs/MaOEAs, or even other categories of multi-objective methods (see Section \ref{sec:MC}). 
 For example, the use of diversity-based MOEAs/MaOEAs in approaches that seek more diversity, as Liu et al.~\cite{Liu2012}.

Another concern in this field is regarding real applications and large-scale clustering problems. Some works, as \cite{garcia2017scaling, Garza2018, Zhu2018_MaOEA} improve the scalability on designing more efficient multi-objective evolutionary algorithms; however, most of the existing multi-objective evolutionary clustering algorithms are not well scalable to real-life applications that generate a huge amount of data. According to Mukhopadhyay et al.~\cite{mukhopadhyay2015survey} it is a challenge for researchers to devise fast, scalable algorithms for multi-objective clustering.

Regarding the final selection, we note that some approaches do not provide a final selection method, providing only an evaluation regarding the clustering process in comparison to other approaches. Thus, the decision-maker has to use another tool to select the best solutions from among these approaches. The choice of which mechanisms to use to select the best solution or set of solutions is also a challenge that requires more studies.

Furthermore, this paper also presented some applications of EMOC and the most relevant related papers that can be useful to researchers that are exploring EMOC for a specific purpose. 

\section*{Acknowledgement}
This work was partially supported by the National Council for Scientific and Technological Development (CNPq), Brazil.

\appendix
\section{Clustering Criteria}\label{appA}
In this section, we present the CVIs applied as objective functions in the literature, as introduced in the Section \ref{sec:obj_fc}. 
We considered a common notation in the equations, where $n$ refers to the number of objects in the dataset $\mathbf{X}$, $\pi$  denotes a partition, $k$ denotes the number of clusters in $\pi$,  $\mathbf{c}_i$ refers to the  $i$th cluster that belongs to $\pi$, $\mathbf{x}_a$ denotes a generic object, $n_i$ denotes the number of objects in $\mathbf{c}_i$, $\mathbf{z}_i$  refers to the centroid of cluster $\mathbf{c}_i$, and $\overline{\mathbf{z}}$ represents the centroid of the dataset. Furthermore, $d(.,.)$ denotes the chosen distance function. 

\subsection{Compactness criteria} 

The \textbf{Average Within Group Sum of Squares} ($AWGSS$) is  computed by the average of the distance between each object in the cluster and its centroid, as present in Eq.~\ref{awgss}. It should  be minimized to obtain compact clusters~\cite{Kirkland2011}.

\begin{equation}\label{awgss}
    AWGSS(\pi) =  \sum_{i=1}^k \frac{\sum_{\mathbf{x}\in \mathbf{c}_i} d(\mathbf{x_a},{\mathbf{z}_i})}{n_i}
\end{equation}

The \textbf{overall Deviation} ($Dev$) is computed as the overall summed distance between data points and their corresponding cluster center, as defined in Eq.~\ref{deviation}. It should be minimized in order to obtain compact clusters~\cite{Handl2005a}. 

\begin{equation}\label{deviation}
Dev(\pi) =  \displaystyle\sum_{\mathbf{c}_i \in \pi}\displaystyle\sum_{\mathbf{x}_a \in \mathbf{c}_i}d(\mathbf{x}_a,\mathbf{z}_i)
\end{equation}

Sert et al.~\cite{Sert2011, sert2012unification} considered the \textbf{K-Mode internal distance} (Km\textsubscript{id}) and \textbf{K-Mode weighted internal distance} (Km\textsubscript{wid}) as objective functions.  These indices are computed in a similar way to $Dev$,  but the mode is used instead of the centroid. Km\textsubscript{id} and Km\textsubscript{wid} should be minimized as objective functions. 

The \textbf{intra-cluster Entropy} ($Ent$) measures the degree of similarity between each cluster center and the data objects that belong to that cluster, as the probability of grouping all the data objects into that particular cluster. A larger value of this index implies better clustering~\cite{Ripon2006_ICPR, Ripon2006_IJCNN, Ripon2009}.  This index is defined by Eq.~\ref{entropy}, where $g(\mathbf{z}_i)$ is the average similarity between $\mathbf{z}_i$ and the data object belong to cluster $\mathbf{c}_i$, and the $cos(.,.)$  represents the cosine distance.

\begin{equation}\label{entropy}
\begin{gathered}
Ent(\pi) = \sum^k_{i=1} \left [(1-h(c_i))g(\mathbf{z}_i) \right ]^{1/k},  \textrm{ where } \\ 
\begin{split}
  &h(c_i) =-[(g(\mathbf{z}_i)  \log_2 g(\mathbf{z}_i)+(1-g(\mathbf{z}_i))  \log_2(1-g(\mathbf{z}_i))],\textrm{ and}  
  &g(\mathbf{z}_i) = \frac{1}{n_i}  \sum^{n_i}_{a=1} \left (0.5 + \frac{\cos(\mathbf{z}_i,\mathbf{x}_a)}{2} \right ) 
 \end{split}
\end{gathered}  
\end{equation}

The \textbf{Homogeneity} ($H$) index is computed by the sum of the average minimal intra-cluster distance, according to Eq.~\ref{homogeneity}, where $\min(d(\mathbf{z}_i, \mathbf{x}_a))$ denotes the lowest distance between the points $\mathbf{x}_a$ in the cluster $\mathbf{c}_i$ and the cluster mode $\mathbf{m}_i$. $H$ should be maximized to obtain homogeneous clusters~\cite{Dutta2012}. 

\begin{equation}\label{homogeneity}
    H(\pi) =\sum\limits^k_{i=1} \left[\frac{\sum^{n_i}_{a=1} \min(d(\mathbf{m}_i, \mathbf{x}_a))}{n_i} \right] 
\end{equation}

The \textbf{intra-cluster Variance} ($Var$) is conceptually similar to $Dev$, as shown in Eq.~\ref{var}, and it also should be minimized to obtain compact clusters~\cite{Garza2018}. 

\begin{equation}\label{var}
	Var(\pi) =  \frac{1}{n} \displaystyle\sum_{\mathbf{c}_i \in \pi}\displaystyle\sum_{\mathbf{x}_a \in \mathbf{c}_i}d(\mathbf{x}_a,\mathbf{z}_i)
\end{equation}

The \textbf{Total Within-Cluster Variance} ($TWCV$) is also applied to identify sets of compact clusters, as defined in Eq.~\ref{TWCV}, where $f$ is the size of the dimensional feature space, $\mathbf{x}_{ar}$ denotes the $r$th feature value of the $a$th data point, $\mathbf{z}_{ir}$ is the centroid of the $i$th cluster of the $r$th feature, and  $w_{ai} \in \left[ 0,1\right]$ and $\sum^k_{i=1} w_{ai}=1$. The goal is to minimize $TWCV$ to obtain compact clusters~\cite{Du2005}.

\begin{equation}\label{TWCV}
\begin{gathered}
    TWCV(\pi) = \sum^k_{i=1}\sum^n_{a=1}  {w}_{ai} \sum^f_{r=1}
    (\mathbf{x}_{ar}-\mathbf{z}_{ir})^2, \textrm{where }\\
    \begin{split}
     &\mathbf{z}_{ir} = \frac{\sum^n_{a=1} w_{ai} \mathbf{x}_{ar}}{\sum^n_{a=1}w_{ai}}, \textrm{ and }
      &w_{ai} \begin{cases}
    {1}, \text{if}\ a^{th} \textrm{object belongs to the } i^{th} cluster\\
    0, \text{otherwise} 
    \end{cases}
    \end{split}
\end{gathered}
\end{equation}

The \textbf{Fuzzy Compactness} ($J_m$) represents the global fuzzy cluster variance, as defined in Eq.~\ref{jm}, where $u_{ia}$ is the membership degree of the $a$th data point to the $i$th cluster, and $m$ is the fuzzy exponent. The smaller value of $J_m$ corresponds to more compact clusters~\cite{bezdek2013pattern}. 

\begin{equation}\label{jm}
J_m =\sum_{i=1}^k\sum_{a=1}^n   u_{ia}^m d(\mathbf{z}_i, \mathbf{x}_a) 
\end{equation}

Zhu et al. introduced an adapted $J_m$ that considers the cluster weighting subspace, the \textbf{Fuzzy weighting subspace clustering} ($J_{wm}$). This index is defined in Eq.~\ref{eq:jwm}, where $f$ is the number of attributes (or vector of features),  $\mathbf{x}_ar$ denotes $r$th feature of the $a$th  object, and $\mathbf{z}_{ir}$ is the centroid of the $i$th cluster of the $r$th feature. $w_{ir}$ is defined in Eq.~\ref{eq:weight}, where  $m$ is the fuzziness exponent, and $\tau$ is the fuzzy weighting index.  $J_{wm}$ should be minimized to improve the clustering~\cite{Zhu2012}.   

\begin{equation}\label{eq:jwm}
    J_{wm} = \sum^{k}_{i=1}\sum^n_{a=1}u^m_{ia}\sum^f_{r=1}w^{\tau}_{ir}d(\mathbf{x}_{ar} - \mathbf{z}_{ir})^2 
\end{equation}

\begin{equation}\label{eq:weight}
\begin{split}
 &w_{ir} =\frac{\left(\sum^n_{a=1}u^m_{ia}d(\mathbf{x}_{ar}-\mathbf{z}_{ir})^2\right)^{1/\tau-1}}{\sum^f_{r=1}\left(\sum^n_{a=1}u^m_{ia}d(\mathbf{x}_{ar}-\mathbf{z}_{ir})^2\right)^{1/\tau-1}},  \textrm{ where }
 u_{ia} = \frac{(\sum^f_{r=1} w_{ia}^\tau d(\mathbf{x}_{ar} - \mathbf{z}_{ir})^2)^{-1/m-1}}{
 \sum^k_{i=1}
 (\sum^f_{r=1} w_{ia}^\tau d(\mathbf{x}_{ar} - \mathbf{z}_{ir})^2)^{-1/m-1}
 }
  \end{split}
\end{equation}

\subsection{Connectedness criteria} 

The \textbf{Connectivity} ($Con$) index~\cite{Handl2005a} evaluates the degree to which neighboring data points have been placed in the same cluster.  This index is computed according to Eq.~\eqref{eq:con}, where $L$ is the parameter that determines the number of nearest neighbors that  contribute to the connectivity, $nn_{ab}$ is the $b$th nearest neighbor of object $\mathbf{x}_a$. $Con$ as objectives should be minimized.

\begin{equation}\label{eq:con}
\begin{gathered}
    Con(\pi) = \sum_{a=1}^{n}\sum_{b=1}^{L}f(\mathbf{x}_a,nn_{ab}),  \textrm{ where } f(\mathbf{x}_a,nn_{ab}) \begin{cases}
    \frac{1}{b},  \text{if}\ \nexists \mathbf{c}_k : \mathbf{x}_a, nn_{ab} \in \mathbf{c}_i \\ 
    0,  \text{otherwise} 
    \end{cases}
    \end{gathered}
\end{equation}

The \textbf{Data Continuity Degree} ($DCD$) measures the connectedness of the data in terms of the connectivity factor (the total edges sum for each minimum spanning tree) in a similarity graph. In general, it can be computed in two steps.  First, a similarity function is applied in order to generate a similarity graph, the $k_{size}$-Graph. In this graph, a vertex $v_a$ is connected with the vertex $v_b$ if $v_b$ is among the $k$-nearest neighbors of $v_a$. After that, the total minimal spanning tree edges are computed considering all nodes connected within the neighborhood of the current node and internally --- this process is repeated with each connected component due to the graph  not being fully connected. The average arithmetic value of the metric (the connectivity factor divided by the number of clusters) is the result of this objective, which should be maximized in the optimization~\cite{menendez2013multi}.

\subsection{Separation criteria} 

The \textbf{Average Between-Group Sum of Squares} ($ABGSS$) is computed as the average distance between the clusters' centroids and the centroid of the data, as defined in Eq.~\ref{abgss}. It should be maximized to obtain well-separated clusters~\cite{Kirkland2011}. 

\begin{equation}\label{abgss}
    ABGSS(\pi) =  \frac{\sum_{i=1}^k n_i. d(\mathbf{z}_i,\overline{\mathbf{z}})}{k}
\end{equation}

The inter-cluster distance \textbf{Average Separation} ($Sep\textsubscript{AL}$)  measures the average separation distance between all clusters, according to Eq.~\ref{sep_al}. $Sep\textsubscript{AL}$ should be maximized to obtain better clustering~\cite{Ripon2009}.

\begin{equation}\label{sep_al}
    Sep_{AL}(\pi) = \frac{1}{k(k-1)/2} \sum^k_{i\neq j} d(\mathbf{z}_i, \mathbf{z}_j), 
\end{equation}

Sert et al.~\cite{Sert2011, sert2012unification} introduce the use of \textbf{K-Mode external distance} ($Km\textsubscript{ed}$) and \textbf{K-Mode weighted external distance} ($Km\textsubscript{wed}$) as objective functions.  These measures are similar to $Sep_{AL}$, however considering the mode instead of the centroid. $Km\textsubscript{ed}$ and $Km\textsubscript{wed}$ should be maximized as objective functions. 

The \textbf{Separation Index} ($Sep\textsubscript{CL}$) is computed by the sum of the distance between every two tuples (data points) in different clusters, according to Eq.~\ref{sep_cl}. It should be maximized to get well-separated clusters~\cite{dutta2012data}.  

\begin{equation}\label{sep_cl}
    Sep_{CL}(\pi) = \sum_{\substack{\mathbf{x}_a \in \mathbf{c}_i,\mathbf{x}_b \in \mathbf{c}_j,\\ i \neq j}} {d(\mathbf{x}_a, \mathbf{x}_b)}
\end{equation}

The \textbf{graph-based separation} index ($Sep_{graph}$) measures the separation between the clusters in terms of  a similarity graph. As in the $DCD$ index, it considers the generation of a $K_{size}$-Graph as the first step in computing this index. The $Sep_{graph}$  is calculated as the arithmetic average value of the edge weights between the different clusters, as defined in Eq.~\ref{sep_graph}, where $\mathbf{c}$ is a cluster, $\mathbf{G}$ is the $K_{size}$-Graph, $\mathbf{v}_a$ is the vertex $a$, and $w_{ab}$ is the edge weight value from node $a$ to node $b$.  $Sep_{graph}$ should be maximized to improve cluster separation~\cite{menendez2013multi}.

\begin{equation}\label{sep_graph}
    Sep_{graph} = \left(\frac{\sum_{\mathbf{v}_a \in \mathbf{G}}\{w_{ab}| \mathbf{v}_a \notin \mathbf{c}\}}{\mathbf{G}-\mathbf{c}}\right) / \mathbf{c}
\end{equation}

The \textbf{Fuzzy Separation} ($Sep_{fuzzy}$) index~\cite{Mukhopadhyay2007icit} measures the inter-cluster fuzzy separation. This index is computed according to Eq.~\ref{f:sep_fuzzy}, where the fuzzy membership is defined by $ \mu$\textsubscript{ij}, $d(\mathbf{z}_j, \mathbf{z}_i) $ is the distance between two centroids $\mathbf{z}_i$ and $\mathbf{z}_j$. To get well-separated clusters, the $Sep_{fuzzy}$ should be maximized.

\begin{equation}\label{f:sep_fuzzy}
\begin{gathered}
    Sep_{fuzzy} = \sum^k_{\substack{i,j=1, \\i \ne j}} \mu_{ij}^m {d(\mathbf{z}_i, \mathbf{z}_j)}, 
    \textrm{ where } \mu_{ij} = 2/ \left( \sum^k_{\substack{l=1,\\ l\neq j}}\left(\frac{d(\mathbf{z}_j, \mathbf{z}_i)}{d(\mathbf{z}_j, \mathbf{z}_l)}\right)^{1/(m-1)}\right) 
\end{gathered}
\end{equation}

The \textbf{Fuzzy Overlap Separation} ($Sep_{nfuzzy}$) considers the combination of the $l$-order overlap and inter-cluster separation, composed of a $t$-normal function $\top$ and t-conorm $\bot$ to formulate the Fuzzy Overlap Separation~\cite{Wikaisuksakul2014, paul2018new}.  $Sep_{nfuzzy}$ is defined in Eq.~\ref{f:sep_nfuzzy}, where $u_{ai}$ is the membership degree of the $a$th data point to the $i$th cluster, $O \bot (\mathbf{u}_a(\mathbf{x}_a),k)$ is the overlapping degree that considers triplets of clusters up to a $k$-tuple of clusters combinations. $Sep_{nfuzzy}$ index measures the isolation of clusters, which is preferred to be large. 

\begin{equation}\label{f:sep_nfuzzy}
    Sep_{nfuzzy} = \frac{1}{n} \sum^n_{a=1} \frac{O \bot (\mathbf{u}_a(\mathbf{x}_a),k)}{\max\limits_{i=1, k}\{u_{ai}\}}
\end{equation}

\subsection{Separation and Compactness criteria}

The \textbf{Categorical Data Clustering with Subjective factors} ($CDCS$) index is computed by the ratio of the intra-cluster cohesion and inter-cluster similarity for the categorical data clustering.  This index is defined by Eq.~\ref{CDCS}, where $\mathbf{A}_r$ is a set of attribute values, ${a}_{r}$ denotes the number of attribute values for the $r$th attribute,  $P(\mathbf{A}_r = {a}_{r}^i|\mathbf{c}_i)$ is the probability of ${a}_{r}^i$ for the $r$th attribute in cluster $\mathbf{c}_i$, $ S(\mathbf{c}_p, \mathbf{c}_q)$ denotes a similarity of two clusters, where $S(\mathbf{c}_p, \mathbf{c}_q) = \prod^{f}_{r=1} \left[\sum^{t_r}_{i} \min{P(\mathbf{A}_{r}=a_r^i|\mathbf{c}_p), P(\mathbf{A}_{r}=a_r^i|\mathbf{c}_q)}+\varepsilon \right]$,
and $\varepsilon$ is a small value in case that each component is 0~\cite{Zhu2018_MaOEA}. 

\begin{equation}\label{CDCS}
\begin{gathered}
    CDCS =  \frac{intra}{inter}, \textrm{ where }\\
    \begin{split}
    &intra =\sum^k_{i=1} \frac{|\mathbf{c}_k|}{n} \sum^f_{r=1} \frac{1}{f}(\max\limits^{n_r}_{i=1}P(\mathbf{A}_r = {a}_{r}^i|\mathbf{c}_i))^3, 
    &inter = \frac{\sum^k_{p=1}\sum^k_{q=1} S(\mathbf{c}_p, \mathbf{c}_q)^{1/f} \cdot |\mathbf{c}_p \cup \mathbf{c}_q|}{(k-1)\cdot n}
    \end{split}
\end{gathered}    
\end{equation}

The \textbf{Calinski-Harabasz} ($CH$) index, also known as the variance ratio criterion, is based on the degree of dispersion between clusters. It can take values in [0, $\infty$] with higher values indicating better clustering. $CH$ is computed by the ratio of the sum of between-cluster dispersion and inter-cluster dispersion for all clusters, as defined in Eq.~\ref{CH}~\cite{Zhu2018}.  

\begin{equation}\label{CH}
    CH(\pi) = \frac{\sum_{i=1}^k n_i \cdot d(\mathbf{z}_i,\overline{\mathbf{z}})} {\sum_{i=1}^k \sum_{\mathbf{x}\in \mathbf{c}_i} d(\mathbf{x},{\mathbf{z}_i})} \frac{ (n-k)}{(k-1)}
\end{equation}

The \textbf{Davies-Bouldin} ($DB$) index is computed as the ratio of the sum of within-cluster scatter to between-cluster separation ($R_i$), as defined in Eq.~\ref{DB}. The minimum value of this $DB$ is zero, with lower values indicating a better clustering~\cite{Tsai2012, Zhu2018, dong2018adaptive, dutta2019automatic}.

\begin{equation}\label{DB}
\begin{gathered}
DB(\pi) = \frac{1}{k}   \sum_{i=1}^k R_i , \textrm{ where } \\
\begin{split}
  &R_i =   \max\limits_{j,j \neq i}\left \{ \frac{S_i+S_j}{d(\mathbf{z}_i, \mathbf{z}_j)} \right \}, \textrm{ and } &S_i=\frac{1}{|n_i|} \sum _{\mathbf{x_a}  \in \mathbf{c}_i} d(\mathbf{x_a},\mathbf{z}_i) 
\end{split}
\end{gathered}
\end{equation}

The \textbf{Dunn}  index is computed as the ratio between the minimum inter-cluster distance ($\delta (\mathbf{c}_i,\mathbf{c}_j)$) to the maximum cluster diameter ($\max _{j\leq i \leq k} {\Delta (\mathbf{c}_i)}$), as defined in Eq. (\ref{Dunn}). It is considered that compact and well-separated clusters have a small diameter and a large distance between them. The Dunn index can take values between zero and infinity, and it should be maximized to obtain a well-separated and compact cluster~\cite{liu2010understanding}. 

\begin{equation}\label{Dunn}
\begin{gathered}
    Dunn(\pi) =  \min\limits_{1\leq i\leq k}\left \{\min\limits_{\substack{1\leq j\leq k, \\ j\neq i}}\left   \{ \frac{\delta (\mathbf{c}_i,\mathbf{c}_j)}{\max\limits_{j\leq i \leq k} {\Delta (\mathbf{c}_i)}}   \right \}  \right \},  \textrm{ where }\\ 
    \begin{split}
    &\delta (\mathbf{c}_i, \mathbf{c}_j) =   \min_{\substack{\mathbf{x}_a \in \mathbf{c}_i,\\ \mathbf{x}_b \in \mathbf{c}_j}}\left \{ d(\mathbf{x}_a,\mathbf{x}_b) \right \}, \textrm{ and } 
    &\Delta (\mathbf{c}_i)=\max_{\mathbf{x}_a,   \mathbf{x}_b \in \mathbf{c}_i} \{d(\mathbf{x}_a,\mathbf{x}_b)\}
    \end{split}
\end{gathered}
\end{equation}

The \textbf{Modularity} ($Mod$) was initially proposed as a measure of the strength of the network's module division. This index is computed as the total difference between the sum of distances of the objects in the same cluster $\mathbf{c}_i$ (that indicates how closely similar data is with others in the same cluster) and the sum of distances considering the objects in the dataset $\mathbf{X}$ (that determines how closely similar data is with others in different clusters), as defined in Eq. \ref{eq:mod}~\cite{Liu2018}.  

\begin{equation}\label{eq:mod}
\begin{gathered}
Mod(\pi) =  \sum^k_{i=1}(cd - od^2), \textrm{ where }\\
\begin{split}
&cd = \frac{\sum_{\substack{ \mathbf{x}_a,\mathbf{x}_b \in \mathbf{c}_i}}d(\mathbf{x}_a,\mathbf{x}_b)}{\sum_{\mathbf{x}_a,\mathbf{x}_b \in \mathbf{X}}d(\mathbf{x}_a,\mathbf{x}_b)}, \textrm{ and }
&od = \frac{ \sum_{\substack{\mathbf{x}_a \in \mathbf{c}_i,\mathbf{x}_b \in \mathbf{X} }}d(\mathbf{x}_a,\mathbf{x}_b)}{\sum_{\mathbf{x}_a,\mathbf{x}_b \in \mathbf{X}}d(\mathbf{x}_a,\mathbf{x}_b)}
\end{split}
\end{gathered}
\end{equation}

The \textbf{Silhouette} ($Sil$) index measures how much each point in the data is similar to its own cluster compared to other clusters, based on the relation of the mean similarity of the objects within a cluster and the mean distance to the objects in the other clusters.  $Sil$ is defined in Eq. \ref{sil}, in which $ad_a$ refers to the mean distance between a sample $\mathbf{x}_a$ and all other points in the same cluster. Moreover, $bd_a$ is the mean distance between a sample $\mathbf{x}_a$ and the nearest cluster that  $\mathbf{x}_a$ is not a part of. Thus, $Sil$ produces values between $-1$ and $1$. A higher value corresponds to a better clustering result~\cite{Mukhopadhyay2007}.

\begin{equation}\label{sil}
\begin{gathered}
Sil(\pi) =   \frac{1}{n} \sum_{a=1}^n S(\mathbf{x}_a), \textrm{where } S(\mathbf{x}_a) =   \frac{bd_a - ad_a}{max\left \{ ad_a,bd_a \right \}}   
\end{gathered}
\end{equation}

The $\mathbb{\mathcal{I}}$ index  measures separation based on the maximum distance between cluster centers, and measures compactness based on the sum of
distances between objects and their cluster centers. This index is computed according to Eq. \ref{eq:iindex}, in which
$E_k$ stands for within cluster scatter, $D_k$ stands for between-cluster separation, $E1$ and $P$ are correlation coefficients, $u_{ia}$ is the membership degree of the $a$th object to the $i$th cluster. A larger value of this index implies better clustering.
~\cite{dong2018adaptive}.

\begin{equation}\label{eq:iindex}
\begin{gathered}
    \mathcal{I} = \left( \frac{1}{k} \cdot \frac{E_1}{E_k} \cdot D_k\right)^P, \textrm{ where } \\ 
    \begin{split}
    &D_k=\max^k_{i,j=1}(\mathbf{z}_i-\mathbf{z}_j),  \textrm{ and }
    &E_k = \sum^k_{i=1}\sum^n_{a=1}u_{ia}(\mathbf{x}_a-\mathbf{z}_i)
    \end{split}
\end{gathered}
\end{equation}

The \textbf{Addition feature weight} ($J_{Add}$) index is applied to minimize both the negative weight entropy and the separation between clusters. This index is defined in Eq.~\ref{J_add}, where $f$ is the number of attributes, and $w_{ir}$ takes the value in [0, 1], which corresponds to a soft partition of features. It is composed by $Sep_i$, that is computed according to Eq. (\ref{f:sep_fuzzy}), $\sigma$ a present value that prevents the denominator from becoming zero, and $A_{wi}$ denotes the average value of the important weights, which are more than or equal to the mean value $(1/f)$ for the $i$th cluster~\cite{xia2013novel}.

\begin{equation} \label{J_add}
\begin{gathered}
    J_{Add} = \sum^k_{i=1} 
    \left( \frac{Aw_i}{(Sep_i+\sigma)}+ \sum^f_{r=1}w_{ir} \log w_{ir} \right), \\
    \begin{split}
      \textrm{ where }&Aw_i = \frac{\sum^f_{r=1}\delta_r w_{ir}}{\sum^f_{r=1}\delta_k}, \textrm{ and } &\delta_r =  
     \begin{cases}
        1,  \text{if}\  w_{ir}>1/f \\
        0,  \text{otherwise}
     \end{cases}
     \end{split}
\end{gathered}
\end{equation}
\noindent

The \textbf{Xeni-Beny} ($XB$) index is defined as a function of the ratio of the total fuzzy cluster variance ($J_m$) to the minimum separation of the clusters ($Sep$), as presented in Eq. \ref{XB}, where $u_{ia}$ is the membership degree of the $a$th data point to the $i$th cluster, and $m$ is the fuzzy exponent. It should be minimized to obtain well-separated and compact clusters~\cite{DiNuovo2007, Zhu2018_MaOEA}. 

\begin{equation}\label{XB}
    XB(\pi) = \frac{J_m   }{n \cdot sep} 
    =\frac{\sum_{i=1}^k\sum_{a=1}^n u_{ia}^m d(\mathbf{z}_i, \mathbf{x}_a)} {n \cdot(\min\limits_{i \neq j} \left\{ d(\mathbf{z}_i, \mathbf{z}_j)\right\} )} 
\end{equation}

The \textbf{Soft Subspace Xie-Beni} ($SSXB$) index was extended from the $XB$, and defined as the ratio of the fuzzy weighting within-cluster compactness ($J_{wm}$) to the fuzzy minimum weighting between-cluster separation ($J_{wsep}$). This index is computed according to Eq.~\ref{SSXB}, and it should be minimized as an objective function~\cite{Zhu2012}. 

\begin{equation}\label{SSXB}
\begin{split}
SSBX(\pi) &= \frac{J_{wm}}{n \cdot J_{wsep}}  =\frac{\sum\limits^k_{i=1}\sum\limits^n_{a=1}u^2_{ia}\sum^f_{r=1}w^\tau_{ir}d(\mathbf{x}_{ar}-\mathbf{z}_{ir})^2}{n\cdot \min_{i \neq j}\{d^2(\mathbf{z}_{ir},\mathbf{z}_{jr})\} } 
\end{split}
\end{equation}

\noindent
where $d^2(\mathbf{z}_{ir},\mathbf{z}_{jr})= (\sum\limits^f_{r=1}w_{ij}^\tau d(\mathbf{z}_{ir}-\mathbf{z}_{jr})^2+\sum\limits^f_{r=1}w_{ij}^\tau d(\mathbf{z}_{ir}-\mathbf{z}_{jr})^2)/2$,  $f$ is the number of attributes. $w_{ir}$ and $u_{ia}$ are defined in Eq.~ \ref{eq:weight}.

\subsection{Other criteria} Here, we present the other criteria applied as objective functions. Cluster cardinality and expected weighted coverage density indices consider the relation  between  the occurrence of objects in a categorical dataset. The similarity index is the only relative CVI used as the objective function, while the other CVIs consider the data properties of each partition. The sparsity and reconstruction error are two particular objective functions designed for spectral clustering.

The \textbf{Cluster Cardinality Index} ($CCI$) considers a set of operations to describe the property and structure of categorical data~\cite{Zhu2018_MaOEA}. It is computed according to Eq.~\ref{CCI}, where $\mathbf{A}_{lr}$ and $\mathbf{A}_{ir}$ are the set of categorical values of $r$th attribute within the clusters $\mathbf{c}_i$ and  $\mathbf{c}_l$. A larger value of CCI implies better clustering.

\begin{equation}\label{CCI}
\begin{gathered}
CCI = \frac{1}{k} \sum^k_{i=1}\max\limits_{l,i=1, l \neq i} \left( \frac{CI(i)+CI(l)}{CI(i,l)}\right),  \textrm{ where }\\
\begin{split}
    &CI(i) = \frac{1}{f} \sum^f_{r=1} \frac{|\mathbf{A}_{ir}|}{\mathbf{c}_i}, \textrm{ and }  &CI(i,l)=\frac{1}{f} \sum^f_{r=1} \frac{|\mathbf{A}_{ir} \cap\mathbf{A}_{lr}| - |\mathbf{A}_{ir} \cup\mathbf{A}_{lr}|+1}{|\mathbf{A}_{ir} \cap\mathbf{A}_{lr}|+1}
\end{split}
 \end{gathered} 
\end{equation}

The \textbf{intra-cluster Expected Weighted Coverage Density} ($EWCD$) considers the relation between the objects in a transational dataset. The transational dataset is composed of $n$ transactions considering  the set of items $\mathbf{I}=\{\mathbf{I}_1, \mathbf{I}_2,\ldots, \mathbf{I}_m\}$, where the transaction $\mathbf{t}_j (1 \leq j \leq n)$ is a set of items  $\mathbf{t}_j=\{\mathbf{I}_{j1}, \mathbf{I}_{j2},\ldots, \mathbf{I}_{jl}\}$, such that $\mathbf{t}_j\subseteq \mathbf{I}$. In this context, the WCD-Weighted Coverage Density  of one cluster is defined as the sum of occurrences of all items in a cluster divided by the number of distinct items and the total number of items in this cluster. Thus, the EWCD of the partition $\pi$ is defined as a average sum of the WCD in all clusters, as presented in the Eq. \ref{eq:ewcd}, where $\mathbf{I}_{ij}$ is the $j$th item set in the cluster $\mathbf{c}_i$, $occur(\mathbf{I}_{ia})$ define the number of occurrences of the  $a$th item in cluster $\mathbf{c}_i$, and ${S}_i$ is the sum occurrences of all items in cluster $\mathbf{c}_i$~\cite{Sert2011, sert2012unification}. 

\begin{equation}\label{eq:ewcd}
\begin{split}
    EWCD (\pi)&= \sum^k_{i=1}\frac{n_i}{n} WCD 
    = \frac{1}{n} \sum_{i=1}^k \left[\frac{\sum_{a=1}^{n_i} occur(\mathbf{I}_{ia})^2}{{S}_i} \right] 
\end{split}    
\end{equation}

Li et al.~\cite{li2017ensemble} introduced the \textbf{Similarity} ($Sim$) index to  evaluate the similarity of one partition to others with a similarity matrix, as defined in Eq.~\ref{Sim}. This index can be used to evaluate the diversity of the solutions in an evolutionary approach. It should be minimized as an objective~\cite{li2017ensemble}.

\begin{equation}\label{Sim}
    Sim = \frac{1}{n} \sum^n_{j=1} similarity({\pi}_i, {\pi}_j), 
\end{equation}

Luo et al.~\cite{Luo2015} modeled the similarity matrix for spectral clustering into objective functions. They assume that $\mathbf{y=Ax}$ is a linear equation of an under-determined system, where $\mathbf{A} \in \mathbb{R}^{M\cdot N}$ is a full-rank and over-complete matrix, which is called an over-complete dictionary, $\mathbf{y} \in \mathbb{R}^{M} $ is called a measurement vector, and $\mathbf{x} \in \mathbb{R}^N$ is a sparse vector. Thus, they use $\mathbf{x}$ and $\mathbf{A}$ to reconstruct $\mathbf{y}$. For that, the \textbf{SParsity} ($SP$), Eq.~\ref{eq:sp}, and \textbf{Reconstruction Error} ($RE$), Eq.~\ref{eq:re}, should be minimized.

\begin{equation}\label{eq:sp}
    SP = \left \|\mathbf{x}\right \|_0,  
\end{equation}

\noindent
where $l_0$ {norm} $\left \|. \right \|_0$ counts the number of nonzero values in a vector.

\begin{equation}\label{eq:re}
\begin{gathered}
    RE = \left \| \mathbf{Ax} -\mathbf{A} \right \|^2_2, 
\end{gathered}    
\end{equation}

\noindent
where  $\left \|. \right \|_2^2$ is the Euclidean norm on signals of a square matrix.

\bibliographystyle{elsarticle-num}
\bibliography{mybibfile}   



\end{document}